\PassOptionsToPackage{dvipsnames,svgnames,table}{xcolor}
\documentclass[runningheads]{llncs}
\usepackage[T1]{fontenc}

\usepackage{graphicx}
\usepackage[american]{babel}

\usepackage{listings, booktabs, array, pifont, amsmath, amssymb, amsfonts, hyperref, multirow, pgfplots}

\pgfplotsset{compat=1.18}
\usepackage{algorithm}
\usepackage{algpseudocode}
\usepackage{stmaryrd}
\usepackage{bbm}

\usepackage{bold-extra}
\newcommand{\set}[1]{\ensuremath{\mathcal{#1}}}

\newcommand{\tool}[1]{\ensuremath{\mathtt{#1}}}

\newcommand{\trajectory}{\ensuremath{\mathbbm{T\!r}}}

\usepackage{tikz}
\usetikzlibrary{matrix, calc, chains, positioning, decorations.pathreplacing, decorations.text, shapes, arrows, fit, backgrounds, calligraphy,matrix}

\usepackage[dvipsnames,svgnames,table]{xcolor}
\usepackage{fontawesome}
\newcommand{\cmark}{\textcolor{ForestGreen}{\faCheck}} 
\newcommand{\xmark}{\textcolor{red}{\faTimes}} 

\definecolor{docugamigreen}{RGB}{75, 191, 68}
\definecolor{grisbleu}{RGB}{56, 66, 87}

\usepackage{circledsteps}

\definecolor{codegray}{rgb}{0.5,0.5,0.5}
\definecolor{codeblue}{rgb}{0.25,0.5,0.75}
\definecolor{codegreen}{rgb}{0,0.6,0}

\lstdefinelanguage{PlainText}{
    basicstyle=\ttfamily\small,
    keywordstyle=\color{docugamigreen}\bfseries,
    commentstyle=\color{codegreen},
    stringstyle=\color{codegray},
    morekeywords=[1]{MODULES,CONTEXT,QUESTION,TABLE,\#END,SOLUTION,SIMPLIFIED,Answer_Generator,Scale_Finder,Solution_Generator,Program_Executor,Program_Generator_And_Verifier,Knowledge_Retrieval,Span_Extractor,Context_Extractor,Column_Lookup,Row_Lookup},
    alsoletter={\#},
}

\usepackage{cite}
\begin{document}

\title{MATATA: Weakly Supervised End-to-End \underline{MA}thematical \underline{T}ool-\underline{A}ugmented Reasoning for \underline{T}abular \underline{A}pplications}

\titlerunning{MATATA: Tool-Augmented Reasoning for Tabular Applications}

\author{
Vishnou Vinayagame\inst{1,2} \and
Gregory Senay\inst{1} \and
Luis Mart\'i\inst{1}
}

\authorrunning{V. Vinayagame, G. Senay, and L. Mart\'i}

\institute{
Docugami Inc, Kirkland, WA, USA.\\
\and
University of Toronto, Toronto, ON, Canada.
}

\maketitle              
\begin{abstract}
Business documents often contain substantial tabular and textual information with numerical values, requiring mathematical reasoning for effective document understanding.
While Small Language Models (SLMs) still struggle at this task, tool-augmented multi-step agents perform better, at the cost of relying on closed-source or larger models, external data, or extensive prompt-engineering.
This work introduces MATATA, a novel weakly supervised end-to-end approach to train multi-step reasoning language agents for document tabular applications.
MATATA presents an annotation-free paradigm for each agent to enhance 3.8B/8B SLMs.
During its two-stage training, MATATA uses the final outcome of the multi-step reasoning chain as weak supervision.
This approach avoids having to individually supervise each intermediate agent in the reasoning chain.
By employing an adaptive planner and shared tools across different datasets, MATATA shows robust performance.
Experiments demonstrate that MATATA achieves state-of-the-art on FinQA, and on TAT-QA among reasoning methods based on open-source SLMs.
Although being SLM-based, MATATA closely matches GPT-4-based frameworks on TabMWP.
This novel weakly supervised approach enables training an end-to-end multi-step reasoning agent without intermediate supervision, supporting future developments of cost-effective powerful agentic systems.


\end{abstract}

\section{Introduction}
Intelligent document understanding is essential across industries such as finance, law, and healthcare, where large volumes of business documents could benefit from automated analysis \cite{zmigrod2024buddiebusinessdocumentdataset}. 
Financial statements, medical reports, and legal filings often contain textual and tabular data, which intelligent systems can process at scale, leveraging machine learning to enhance document understanding.

Recent developments have demonstrated the potential of Large Language Models (LLMs) \cite{abdin2024phi3technicalreporthighly,touvron2023llama2openfoundation,brown2020languagemodelsfewshotlearners} in tackling those problems \cite{fang2024largelanguagemodelsllmstabular}, though they have yet to reach human-level performance.
In fact, complex reasoning tasks involving multiple steps still pose significant challenges in business document understanding, as measured by multiple benchmarks \cite{chen2021finqa,zhu2021tatqaquestionansweringbenchmark,Zhu_2022}.
Understanding those documents typically involves performing intricate column transformations, querying large and complex tables or computing numerical calculations. 
Therefore, tool-augmented LLMs \cite{gou2024toratoolintegratedreasoningagent,kim2024huskyunifiedopensourcelanguage,lu2023chameleonplugandplaycompositionalreasoning,yin2024mumathcodecombiningtooluselarge} have emerged as a particularly effective approach for complex document understanding.
These frameworks leverage the planning capabilities of LLMs to orchestrate tools and combine program assistance, data manipulation, natural language reasoning, and many others, enabling multi-step problem-solving.
Closed-source and larger language models can be easily adapted into powerful multi-step reasoning agents through prompt engineering, as demonstrated with the use of ChatGPT and GPT-4 in \cite{srivastava2024evaluatingllmsmathematicalreasoning,lu2023chameleonplugandplaycompositionalreasoning,wei2023chainofthoughtpromptingelicitsreasoning,chen2023programthoughtspromptingdisentangling}.
On the other hand, Small Language Models (SLMs) require more adaptation than prompting due to their inherently limited capabilities, as discussed in \cite{bi2024enhancingreasoningcapabilitiessmall}.
Tool-augmented SLM frameworks usually undergo a teacher-student training approach, which requires extensive data annotation. 
\tikzstyle{every pin edge}=[<-, shorten >=8pt, shorten <=8pt]
\tikzstyle{base}=[rectangle, draw, thick, color=black, fill=black!25, inner sep=2pt, rounded corners=0.05cm, minimum height=0.5cm]
\tikzstyle{tool node}=[base, fill=green!47, text width=0.56cm, text centered, rounded corners=0.05cm]
\tikzstyle{llm node}=[base, fill=gray!29, text centered, rounded corners=0.05cm]
\tikzstyle{process node}=[base, fill=cyan!47, text width=1.91cm, text centered]
\tikzstyle{input node}=[base, fill=orange!47]
\tikzstyle{finetune node}=[tool node, left color=gray!47, right color=magenta!47, text width=1.2cm]
\tikzstyle{kto node}=[tool node, left color=magenta!47, right color=red!47, text width=1.46cm]
\tikzstyle{matata node}=[tool node, fill=magenta!47]
\tikzstyle{tool group}=[rectangle, line width=0.38pt, inner sep=0.15cm, text centered, fill=green!11, rounded corners=0.11cm]
\tikzstyle{ellipsis}=[inner sep=0pt, outer sep=0pt, text width=0.56cm, text centered, draw=none, fill=none]
\tikzstyle{dataset}=[cylinder, draw=black, thick, text=black, 
                     cylinder uses custom fill,
                     cylinder body fill = pink!11, 
                     cylinder end fill = pink!47, 
                     aspect = 0.11, 
                     shape border rotate = 90,
                     text centered,
                     text width = 0.83cm,
                     minimum height=0.74cm, 
                     minimum width=0.92cm]

\tikzstyle{text node}=[base, fill=pink!11]
\tikzstyle{arrow}=[->, thick, shorten >=1.5pt, shorten <=1.5pt]
\tikzstyle{data-arrow} = [arrow, color=pink!64!black]
\begin{figure}[tb]%
    \resizebox{\linewidth}{!}{\def\horizsep{0.92cm}
\def\vertsep{0.74cm}

\newsavebox{\questiontable}
\sbox{\questiontable}{
    \begin{tabular}{lcccccc}
        \toprule
        \multicolumn{7}{c}{\textbf{Fiscal years ended July 31,}} \tabularnewline
        \midrule
        & \multicolumn{2}{c}{\textbf{2019}} & \multicolumn{2}{c}{\textbf{2018}} & \multicolumn{2}{c}{\textbf{Change}} \tabularnewline
        \midrule
        & \textbf{Amount} & \textbf{\% of total revenue} & \textbf{Amount} & \textbf{\% of total revenue} & \textbf{(\$)} & \textbf{(\%)} \tabularnewline
        \midrule
        \multicolumn{7}{c}{(In thousands, except percentages)} \tabularnewline
        \midrule
        Cost of revenue: & & & & & & \tabularnewline
        License and subscription & \$64,798 & 9\% & \$35,452 & 5\% & 29,346 & 83\% \tabularnewline
        Maintenance & 16,499 & 2\% & 14,783 & 2\% & 1,716 & 12\% \tabularnewline
        Services & \textbf{243,053} & 34\% & \textbf{246,548} & 38\% & (3,495) & (1\%) \tabularnewline
        Total cost of revenue & \$324,350 & 45\% & 296,783 & 45\% & 27,567 & 9\% \tabularnewline
        \midrule
        \multicolumn{7}{c}{Includes stock-based compensation of:} \tabularnewline
        \midrule
        Cost of license and subscription revenue & \$3,011 & & \$1,002 & & 2,009 & \tabularnewline
        Cost of maintenance revenue & 1,820 & & 1,886 & & (66) & \tabularnewline
        Cost of services revenue & 22,781 & & 21,856 & & 925 & \tabularnewline
        Total & \$27,612 & & \$24,744 & & 2,868 & \tabularnewline
        \bottomrule
    \end{tabular}
  }
\newsavebox{\rowselectiontable}
\sbox{\rowselectiontable}{
        \begin{tabular}{lcc}
            \toprule
            \multicolumn{3}{c}{\textbf{Fiscal years ended July 31,}} \\
            \midrule
            & \textbf{2019} & \textbf{2018} \\
            \midrule
            & \textbf{Amount} & \textbf{Amount} \\
            \midrule
            \multicolumn{3}{c}{(In thousands, except percentages)} \\
            \midrule
            Cost of revenue: & & \\
            Services & \textbf{243,053} & \textbf{246,548} \\
            \bottomrule
        \end{tabular}
}
\begin{tikzpicture}[node distance=\vertsep and \horizsep,
                    line join=round,
                    line cap=round
]
\newcommand{\questioncolor}{blue}
\newcommand{\contextcolor}{ProcessBlue}
\newcommand{\tablecolor}{SkyBlue}

\begin{scope}[local bounding box=input-texts]
    \node[text node, fill=\contextcolor!29, text width=9cm, scale=0.65, label={Context}] (input-context) {
    Excluding the impact of this reclassification, third-party consultants billable to customers primarily for \textit{InsuranceNow} implementation engagements increased by \$3.2 million and personnel expenses related to new and existing employees increased by \$2.8 million.
    We had \textbf{781} professional service employees and 198 technical support and licensing operations employees at July 31, 2019, compared to \textbf{838} professional services employees and 121 technical support and licensing operations employees at July 31, 2018.
    };
    \node[text node, fill=\questioncolor!29, text width=8cm, scale=0.74, below=of input-context, label={Question}] (input-question) {
    What was the weighted average Services per total service employees for 2018 and 2019?
    
    Provide the scale if there is one.};
    \node[text node, fill=\tablecolor!29, right=0.65*\horizsep of input-question.south east, anchor=south west, label={Table}] (input-table) {\scalebox{0.38}{\usebox{\questiontable}}};
\end{scope}

\matrix [matrix, tool group, fill=orange!11, column sep=0.47*\horizsep, label={left: \texttt{input}:}, below=of input-question] (input) {
    \node[input node, fill=\questioncolor!38] (q) {$\vec{q}$}; &
    \node[input node, fill=\contextcolor!38] (c) {$\vec{c}$}; &
    \node[input node, fill=\tablecolor!38] (t) {$\vec{t}$};\\
};

\draw[arrow,rounded corners] (input-question.south) -- ($(input-question.south) + (0pt, -5pt) $)-- ($(q.north) + (0, 10pt)$) |- (q.north);
\draw[arrow,rounded corners] (input-context.east) -| ($(input-question.south east) + (5pt, -5pt) $) -- ($(c.north) + (0, 10pt)$) |- (c.north);
\draw[arrow,rounded corners] (input-table.south) -- ($(input-table.south) + (0pt, -5pt) $) -- ($(t.north) + (0, 8pt)$) |- (t.north);

\node[matata node, right=3.5*\horizsep of input, text width=1.9cm, minimum height=0.74cm] (planner) {\tool{Planner}};

\tikzstyle{matata tool}=[matata node, text width=1.82cm, minimum height=0.92cm]
\tikzstyle{intermediate result}=[text node, text width=1.82cm]

\matrix [matrix, tool group, draw=magenta, fill=magenta!11, column sep=0.29*\horizsep, below=4*\vertsep of input-texts, label={above: \trajectory, trajectory created by \texttt{Planner} (tool execution plan)}] (trajectory) {
    \node[matata tool, text width=4cm] (t1) {\texttt{Row/Column\\Lookup}};&%
    \node[matata tool] (t2) {\texttt{Context Extraction}};&%
    \node[matata tool] (t3) {\texttt{Program Generator}};&%
    \node[matata tool, text width=1.5cm] (t4) {\texttt{Program Executor}};&%
    \node[matata tool, text width=1.5cm] (t5) {\texttt{Scale Finder}};&
    \node[matata tool] (t6) {\texttt{Answer Extractor}};\\
};

\node[intermediate result, below=1.46*\vertsep of t1,text width=4cm] (res1) {\resizebox{\textwidth}{!}{\usebox{\rowselectiontable}}};
\node[intermediate result, text width=2.63cm, scale=0.74, below=1.46*\vertsep of t2] (res2) {
\textbf{781} professional
service
employees and
[\ldots] July 31, 2019,
compared to \textbf{838}
professional
services
employees and
[\ldots] July 31, 2018.
};
\node[intermediate result, text width=2.63cm, scale=0.74, below=1.46*\vertsep of t3] (res3) {
\texttt{ans = ((781 * 246,548) + (838 * 243,053)) / (781 + 838)}
};
\node[intermediate result, text width=1.5cm, text centered, below=1.46*\vertsep of t4] (res4) {
\texttt{244,738.8}
};
\node[intermediate result, text width=2.2cm, scale=0.74, text centered, below=1.46*\vertsep of t5] (res5) {The answer scale is thousand from the table.};
\node[intermediate result, fill=green!47, below=3*\vertsep of t6, text centered, label={[text width=1.4cm, text centered]below: }] (res6) {Answer is $\mathbf{\$\,244,738.8}$ thousand.};

\begin{scope}[on background layer]
    \node[tool group, fill=orange!11, draw=none, fit=(res1) (res5), label={[label distance=-0.65cm]300:Intermediate results}] {};
\end{scope}

\draw[arrow] (input) -- node[midway, anchor=south] {\tiny \Circled{1} \texttt{input} to \texttt{Planner}} (planner); 
\draw[arrow, rounded corners] (planner.east) node[anchor=south west] {\tiny \Circled{2} tool trajectory, $\trajectory$} -- ($(planner.east) + (2*\horizsep,0)$)  -- ($(trajectory.north) + (5*\horizsep,0)$);

\draw[arrow, rounded corners] (input.south) -- ($(input.south) + (0pt, -5pt) $) -- node[pos=0.83, sloped, anchor=south west] {\tiny \Circled{3} \texttt{input} to $\trajectory$} ($(t1.north west) + (-10pt,10pt)$) |- ($(t1.west) + (-10pt,0)$) -- (t1.west);

\foreach \target in {1,...,5}
    \draw[arrow] (t\target) -- (res\target);

\draw[arrow] (t6) -- node[pos=0.56,rotate=90, anchor=north west, below] {\tiny \Circled{5} final answer} (res6);

\draw[arrow] (res1.north east) to[out=74,in=245] (t2);
\draw[arrow] (res2.north east) to[out=74,in=245] (t3);
\draw[arrow] (res3.north east) to[out=74,in=245] (t4);
\draw[arrow] (res4.north east) to[out=74,in=245] (t5);
\draw[arrow] (res5.north east) to[out=74,in=245] (t6);

\begin{scope}[on background layer]
\coordinate (a) at ($(trajectory.south west) - (-42pt, 0.47*\vertsep)$);
\coordinate (b) at ($(trajectory.south east) - (42pt, 0.47*\vertsep)$);
\node[single arrow, rounded corners=0.029cm, draw=none, fill=Periwinkle!29, 
      minimum width = 11pt, single arrow head extend=2pt,
      inner sep=2pt,
      fit=(a) (b),
      minimum height=4.7mm, label={[xshift=1mm]left:\tiny \Circled{4} execute \trajectory}] {}; 
\end{scope}

\end{tikzpicture}}%
    \caption{Schematic representation of MATATA solving a problem instance $\left\langle\vec{q},\vec{c},\vec{t}\right\rangle$. \scalebox{0.74}{\Circled{1}}~Input is first sent to the MATATA \tool{Planner}. \scalebox{0.74}{\Circled{2}} \tool{Planner} then generates a tool use trajectory, \trajectory, that expresses the sequence of how the tools available to MATATA should be applied to solve the problem. After that, \scalebox{0.74}{\Circled{3}} input is sent to the first tool in \trajectory, and \scalebox{0.74}{\Circled{5}} the sequence of tools is applied to \scalebox{0.74}{\Circled{5}} produce the output of MATATA. 
    }%
    \label{fig:matata-inference}%
\end{figure}

\newcolumntype{P}[1]{>{\centering\arraybackslash}p{#1}}
\newcolumntype{M}[1]{>{\raggedright\arraybackslash}p{#1}}
\newcolumntype{L}[1]{>{\raggedleft\arraybackslash}m{#1}}

\renewcommand{\arraystretch}{1.3} 
\begin{table}[tb]  
    \centering
    \caption{Overview of LLM reasoning frameworks over tabular data, with their learning method, Prompt Engineering (PE) or Fine-Tuning (FT), if they are multi-step, their reliance on GPT-4, their use of human experts and their evaluation datasets.}
    \label{table:frameworks}
    \resizebox{1\textwidth}{!}{
        \begin{tabular}{@{}l@{}cc@{\quad}c@{\quad}c@{\quad}M{0.25\textwidth}@{}}
            \toprule
                               &  & \textbf{Multi-} & \textbf{Uses} & \textbf{Human} & \\[-0.56em] 
            \textbf{Framework} & \textbf{Method} & \textbf{step} & \textbf{GPT-4} & \textbf{experts} & \textbf{Datasets used}\\
            \midrule
            PoT \cite{chen2023programthoughtspromptingdisentangling} & PE & \cmark & \cmark & \xmark & FinQA, TAT-QA, TabMWP\\
            EEDP \cite{srivastava2024evaluatingllmsmathematicalreasoning} & PE & \cmark & \cmark  & \xmark & FinQA, TAT-QA, TabMWP\\
            Chameleon \cite{lu2023chameleonplugandplaycompositionalreasoning} & PE & \cmark & \cmark & \xmark & TabMWP\\
            ToRA \cite{gou2024toratoolintegratedreasoningagent}& FT & \cmark & \cmark & \xmark & TabMWP\\
            Husky \cite{kim2024huskyunifiedopensourcelanguage}& FT & \cmark & \cmark & \xmark & FinQA, TabMWP\\
            TAT-LLM \cite{zhu2024tatllmspecializedlanguagemodel}& FT & \cmark & \xmark & \cmark & FinQA, TAT-QA\\
            FinQANet \cite{chen2021finqa}& FT & \xmark & \xmark & \cmark & FinQA\\
            TagOP \cite{zhu2021tatqaquestionansweringbenchmark}& FT & \xmark & \xmark & \cmark & TAT-QA\\
            \midrule
            \textbf{MATATA (ours)} & FT & \cmark & \xmark & \xmark & FinQA, TAT-QA, TabMWP\\
            \bottomrule
        \end{tabular}
    }
\end{table}

There are two main existing approaches regarding annotations: either they are (i) automatically generated using GPT-4, as in Husky~\cite{kim2024huskyunifiedopensourcelanguage} and ToRA~\cite{gou2024toratoolintegratedreasoningagent}, or (ii) extensively annotated by domain experts, as for TAT-LLM~\cite{zhu2024tatllmspecializedlanguagemodel}.
Although, developing a robust teacher model for each tool demands labor-intensive prompt engineering, involving manual efforts \cite{lu2023chameleonplugandplaycompositionalreasoning} that may not capture the complexity and diversity of the task.
Moreover, using teacher models for data generation can be data inefficient \cite{hosseini2024vstartrainingverifiersselftaught}, often resulting in discarding incorrect solutions, while human expert annotation \cite{zhu2024tatllmspecializedlanguagemodel} is a costly and time-consuming alternative.

As detailed in Table \ref{table:frameworks}, existing tabular data understanding methods navigate a trade-off between multi-step reasoning, reliance on closed-source models, or dependence on human experts. 
Motivated by those challenges, this work introduces the MAthematical Tool-Augmented reasoning for Tabular Applications (MATATA) framework.
MATATA is a novel weak supervision method for end-to-end training of tool-augmented multi-step reasoning frameworks.
MATATA develops specialized models as tools during a two-stage training, through instruction tuning \cite{zhang2024instructiontuninglargelanguage} then preference optimization \cite{ethayarajh2024kto} to effectively enhance multi-step reasoning capabilities without any intermediate supervision.
Fig.~\ref{fig:matata-inference} illustrates how MATATA handles a complex reasoning problem from the TAT-QA \cite{zhu2021tatqaquestionansweringbenchmark} dataset.

The proposed method aims to reduce both prompt engineering efforts as well as reliance on external annotations and closed-source models while achieving comparable or higher performance to the current state-of-the-art methods.

The main contributions of this work are:
\begin{enumerate}
   \item a novel end-to-end weak supervision method to train tool-augmented reasoning frameworks with data-efficiency, and
   \item state-of-the-art performance on tabular document reasoning tasks, achieved through a unified framework that enables efficient training of tools across multiple datasets.
\end{enumerate}
Extensive experiments are conducted with different families of SLMs on 3 diverse tabular and textual document benchmarks: TAT-QA \cite{zhu2021tatqaquestionansweringbenchmark}, FinQA \cite{chen2021finqa} and TabMWP \cite{lu2023dynamicpromptlearningpolicy}.
The rest of this work is organized as follows.
Section~\ref{sec:related} presents the related works.
Section~\ref{sec:matata} introduces MATATA. 
The experimental protocol and the results are shown in Section \ref{sec:experiments}.
Finally, Section \ref{sec:conclu} concludes and discusses directions for further research.

\section{Related Work}\label{sec:related}
Understanding business documents with rich tabular and textual content has gained increasing attention.
One of the most common approaches to such problems is multi-step reasoning over tabular data with numerical or logical computations and complex information extraction from text.
The following sections outline the different methodologies applied to tackle these reasoning tasks effectively.

\subsection{Training-free Multi-step Reasoning Frameworks}
Training-free methods rely on the inherent reasoning abilities of LLMs.
Frameworks such as EEDP~\cite{srivastava2024evaluatingllmsmathematicalreasoning} or Program-of-Thoughts~\cite{chen2023programthoughtspromptingdisentangling} use few-shot prompting for multi-step reasoning.
While effective when relying on GPT-4 or GPT-3.5-Turbo, these strategies fail to generalize when coupled with smaller open-source models such as Llama 2-13B or Mistral-7B, with a performance drop of over 30\% on FinQA and TAT-QA (\emph{e.\,g.} EEDP~\cite{srivastava2024evaluatingllmsmathematicalreasoning}).

Chameleon \cite{lu2023chameleonplugandplaycompositionalreasoning} leverages tool-augmentation, where an LLM-based planner dynamically orchestrates tools for specific tasks like web search, table transformations or Python coding.
By few-shot prompting both the tools and the planner, Chameleon-GPT-4 achieves state-of-the-art performance on TabMWP.
However, as observed in Husky's comparative study \cite{kim2024huskyunifiedopensourcelanguage}, Chameleon's reliance on closed-source models limits its generalization to smaller open-source alternatives, as Chameleon-Tulu2-7B suffers from a 20\% accuracy drop on TabMWP.

Although prompt-engineering strategies minimize the need for training, these approaches are highly sensitive to the prompt design and often require extensive efforts to optimize the instructions and examples for in-context learning, as discussed in \cite{lu2023dynamicpromptlearningpolicy}.

\subsection{Fine-tuned Multi-step Reasoning Frameworks}

While prompt-engineering provides a training-free method, its generalization to smaller open-source models for multi-step reasoning remains limited. 
This highlights the need for computationally reasonable, yet efficient models capable of powering reasoning frameworks to handle intricate document understanding.
Fine-tuning dedicated LLMs for multi-step reasoning is a promising approach, particularly when deploying closed-source models is impractical due to prompt-engineering efforts, cost, latency or data privacy concerns, which can arise when dealing with sensitive business documents as discussed in \cite{zhu2024tatllmspecializedlanguagemodel}.

For instance, TAT-LLM \cite{zhu2024tatllmspecializedlanguagemodel} introduces a stepwise pipeline to break down problems into extraction, reasoning, and execution steps. This decomposition helps open-source LLMs from 7B to 70B fine-tuned on data annotated by human experts to achieve discrete reasoning over tabular financial documents.
Similarly, Husky \cite{kim2024huskyunifiedopensourcelanguage} fine-tunes LLaMA-2 and LLaMA-3 models from 7B to 13B, and ToRA \cite{gou2024toratoolintegratedreasoningagent} fine-tunes LLaMA-2 and CodeLLaMA models up to 70B, equipping them with tool-augmented multi-step reasoning abilities.

Husky incorporates specialized tools for code generation and execution, web search, and commonsense knowledge, while ToRA focuses on dynamical planning with code-based tools leveraging libraries like SymPy \cite{10.7717/peerj-cs.103} and solvers.
Both Husky and ToRA demonstrate that fine-tuning the tools and the planner is crucial to enhance SLMs' reasoning abilities, with extensive comparison studies against other LLM frameworks on multiple reasoning tasks.

\subsubsection{Instruction Tuning} Instruction tuning is typically the first step in fine-tuning language models \cite{zhang2024instructiontuninglargelanguage}, where they are trained on input-output pairs. 
When dealing with multi-step reasoning frameworks, this involves fine-grained supervision, to train on intermediate reasoning steps or tool-specific for tool-augmented frameworks.

To obtain the training data, one approach is to query a teacher model, typically a few-shot prompted closed-source LLM like GPT-4 in both Husky and ToRA
, to solve problems and retain the correct multi-step reasoning data.
Alternatively, in TAT-LLM 
manual annotations from human domain-experts are used to supervise each step of the reasoning pipeline.

Both strategies come with limitations.
Relying on closed-source teacher models brings back the prompt-engineering burden, but can also be costly and raises data privacy concerns \cite{zhu2024tatllmspecializedlanguagemodel}, especially with sensitive business documents.
On the other hand, human annotation is costly and time-consuming, creating a scalability bottleneck for fine-tuning reasoning frameworks on large and diverse datasets.

\subsubsection{Preference Optimization} Preference optimization aims to further align fine-tuned models on complex reasoning tasks.
Traditional approaches like Reinforcement Learning from Human Feedback (RLHF) \cite{ouyang2022traininglanguagemodelsfollow} require a resource-heavy reward model with human annotations.
Direct Preference Optimization (DPO) \cite{rafailov2023direct} offers an efficient alternative by directly optimizing models based on preferred and dispreferred samples.
As shown in Weak-to-Strong Reasoning \cite{yang2024weaktostrongreasoning} and V-STaR \cite{hosseini2024vstartrainingverifiersselftaught} it still requires multiple samples with preference labels for each input step.

More recently, Kahneman--Tversky Optimization (KTO) \cite{ethayarajh2024kto} has emerged as a data-efficient alternative, as it relies on binary preference signals (correct or incorrect) for a given pair of input and output.
This eliminates the need for both preferred and dispreferred samples, or additional reward models.

\subsection{Datasets for Tabular Mathematical Reasoning in Documents}

Developing robust mathematical reasoning frameworks over documents requires high-quality datasets with real-world complexity.
This challenge has become prominent with benchmarks like TAT-QA \cite{zhu2021tatqaquestionansweringbenchmark}, FinQA \cite{chen2021finqa}, and TabMWP \cite{lu2023dynamicpromptlearningpolicy}.
Unlike standard Document Question Answering problems such as DocVQA \cite{mathew2021docvqadatasetvqadocument} or Docmatix \cite{laurençon2024building}, these datasets require multi-step reasoning over tabular and textual data with numerical or logical computations, and complex information extraction.

FinQA and TAT-QA focus on mathematical reasoning over semi-structured financial documents, such as financial reports and earnings releases.
They require table manipulation, numerical computation and information extraction from both tabular data and long textual contexts.
TabMWP targets math word problems in tabular contexts, involving logical reasoning and numerical computations in multiple domains with varying levels of difficulties, from Grade 1 to 8.
Therefore, these datasets serve as essential evaluation grounds to assess the capabilities of multi-step reasoning frameworks.

The methods presented in Table~\ref{table:frameworks} leverage either TAT-QA, FinQA, and TabMWP for training and/or evaluation.
To compare our proposal with the state-of-the-art methods in tabular document understanding, these three datasets have been selected in the experimental protocol.

\section{MATATA: MAthematical Tool-Augmented reasoning for Tabular Applications}\label{sec:matata}

MATATA is an end-to-end weakly supervised method for training LLM tool-augmented reasoning frameworks.
MATATA is meant to answer the following challenges: 
\begin{itemize}
\item Can a tool-augmented reasoning framework be trained end-to-end with weak supervision?
\item Can high-quality reasoning be achieved while reducing reliance on external models, data, and extensive prompt engineering?
\end{itemize}

Different methods and frameworks have already been proposed to answer those challenges (\emph{e.\,g.} Table~\ref{table:frameworks}).
However, their underlying limitation lies in the amount of fine-grained annotated reasoning data needed to train these tools, which is usually not available.
Therefore, those methods rely on external proxies, such as GPT-4 or human annotators, to generate those fine-grained training data, or call closed-source models with extensive prompt engineering. 
The assumption behind end-to-end weak supervision, is that it could provide sufficient control over each tool to effectively train reasoning frameworks without intermediary annotations.

MATATA takes inspiration from Husky
, ToRA
, and Chameleon
.
It employs a planner with specialized tools to generate multi-step reasoning. 
The planner is tasked with generating a reasoning trajectory that break down complex problems into step-by-step solutions, and where each step consists of the execution of a specific tool, from the set of tools available to the planner.

Unlike other existing fine-tuned tool-augmented LLM approaches like Husky or ToRA, MATATA does not rely on GPT-4 to generate fine-grained reasoning training data.
Instead, in MATATA, a language model goes through all the training stages and is fine-tuned on its own generated reasoning data, following the concept of self-improvement put forward in  \cite{huang2022largelanguagemodelsselfimprove}.

On the other hand, the core proposal of methods like Chameleon, Program-of-Thoughts, or EEDP
depends on high-quality prompt-engineering, with optimized instructions and in-context-learning examples used to query closed-source LLMs.
While these approaches yield strong performances, they require extensive hand-crafting efforts.
By contrast, MATATA does not rely on closed-source models and minimizes human involvement in prompt engineering.
This could demonstrate the system's ability to improve with minimal initial guidance, with more details in Section~\ref{sec:experiments}.

Moreover, combining instruction tuning with KTO preference optimization is particularly suited for training multi-step reasoning frameworks in a weak-supervised manner, where the expected final answer can be used as signal to supervise each intermediate reasoning step.
Based on this, MATATA suggests a novel way of fine-tuning tool-augmented frameworks, without relying on expensive human annotations, heavy prompt engineering, or closed-source models.

\subsection{MATATA Framework}

\begin{definition}[Tabular Reasoning Problem] \label{def1}
Given a question $\vec{q}$, a context $\vec{c}$, a table $\vec{t}$, and an expected answer $\vec{a}$; find a model $\mathcal{M}\left(\right)$ such that,
\begin{equation}
\mathcal{M}\left(\vec{q},\vec{c},\vec{t}\right)=\hat{\vec{a}}\stackrel{\scriptstyle\boldsymbol{\cdot}}{=}\vec{a}\,,
\end{equation}
where $\hat{\vec{a}}$ represents the predicted answer.
\end{definition}

Using the above notation, a tabular reasoning dataset can be expressed as a set of tuples of question, context, table and expected answer for each document $d$,
\begin{equation}
\set{D} = \left\{\left\langle \vec{q}_d,\vec{c}_d,\vec{t}_d, \vec{a}_d \right\rangle\ \mid \forall d \in \llbracket 1, D \rrbracket\,\right\}.
\end{equation}

\begin{definition}[MATATA Model] \label{def2}
A MATATA model instance 
\begin{equation}
    \mathcal{M}\left(\vec{q},\vec{c},\vec{t} \mid \tool{P},\left\{\tool{T}_i\right\}_{i \in \llbracket 1, T \rrbracket}\right)\,,
\end{equation} 
consists of a planner $\tool{P}$, and a set of tools $\left\{\tool{T}_i\right\}_{i \in \llbracket 1, T \rrbracket}$ that contains the atomic action problem-solving tools available. See Fig.~\ref{fig:matata-inference} for an illustrative example of MATATA on a tabular reasoning application.
\end{definition}

Each tool $\tool{T}_i$, transforms (enhances) the current question, context, table, and answer into an result according to its role, either adding or filtering information:
\begin{equation}
    \tool{T}_i\left(\vec{q},\vec{c_{i-1}},\vec{t_{i-1}},\vec{a_{i-1}}\right)= \left(\vec{q},\vec{c_i},\vec{t_i},\vec{a_i}\right)\,.
\end{equation}
Similarly, the planner $\tool{P}$, determines a \emph{reasoning trajectory}, $\trajectory$, which defines the sequence of tool calls to be executed in order to generate a solution. Hence, for each document $d$,
\begin{equation}
\tool{P}\left(\vec{q_d},\vec{c_d},\vec{t_d}\right) \rightarrow {\trajectory}_{d} := \left\langle\tool{T}_{\text{start}},\ldots,\tool{T}_{i},\ldots\tool{T}_{\text{end}}\right\rangle_d\,.
\end{equation}
Therefore, a final predicted answer, $\hat{\vec{a}}$, is generated by the composition of the tool calls in the trajectory,
\begin{equation}
    \left(\varnothing,\varnothing,\varnothing,\hat{\vec{a}}\right) = \tool{T}_{\text{end}} \circ \ldots \circ \tool{T}_{i} \ldots \circ \tool{T}_{\text{start}}\left(\vec{q},\vec{c},\vec{t}, \varnothing\right)\,.
\end{equation}

As MATATA's focus is on tabular mathematical reasoning, the selected tools include tabular data manipulation, code generation, or information augmentation to solve those complex problems.
The planner is provided with the available tools and their description to generate the reasoning trajectory.
Those tools are described in Table~\ref{tab:tool-description}.

\begin{table}[tb]
    \centering
    \caption{Tools available to MATATA in the experiments of this work.}
    \label{tab:tool-description}
    \begin{tabular}{@{}p{0.29\textwidth}@{}p{0.7\textwidth}@{}}
    \toprule
        \textbf{Tool name} & \textbf{Description} \\
    \midrule
        \textcolor{docugamigreen}{\texttt{Row\_/Column\_Extractor}} & Selects rows and/or columns from large tables.\\
        \textcolor{docugamigreen}{\texttt{Context\_Extractor}} & Retrieves meaningful information from large textual contexts in the documents.\\
        \textcolor{docugamigreen}{\texttt{Span\_Extractor}} & Extracts lists of text spans from the documents.\\
        \textcolor{docugamigreen}{\texttt{Knowledge\ Retrieval}} & Verbalizes knowledge on domain-specific questions.\\
    \midrule
        \textcolor{docugamigreen}{\texttt{Program\_Generator}} & Generates code for numerical computation.\\
        \textcolor{docugamigreen}{\texttt{Program\_Executor}} & Executes code and extract its outputs.\\
    \midrule
        \textcolor{docugamigreen}{\texttt{Solution\_Generator}} & General purpose natural language reasoning.\\
        \textcolor{docugamigreen}{\texttt{Scale\_Finder}} & Finds the scale of the answer when there is one.\\
        \textcolor{docugamigreen}{\texttt{Answer\_Extractor}} & Parses and extracts the final answer of the model.\\
    \bottomrule
    \end{tabular}
\end{table}

\subsection{MATATA Training}\label{sec:matata-training-phases}

MATATA training is structured in four phases that successively take place to progressively generate training data, fine-tune models with instruction tuning, generate improved training data, and fine-tune models with preference optimization. This process is summarized in schematic form in Fig.~\ref{fig:matata-training}.

\begin{figure}[tb]
    \centering
    \resizebox{\linewidth}{!}{
    \input{matata-training-tikz-v2}
    }
    \caption{Schematic representation of MATATA training phases. Phase 1 relies on a prompt-engineered MATATA ($\mathcal{M}^{\text{PE}}$) to generate a training dataset $\set{D}^\text{PE}$ containing tool-augmented reasoning trajectories ($\trajectory_d$) that yield correct answers. Phase 2 uses $\set{D}^\text{PE}$ for a weakly supervised LoRA instruction tuning over a pretrained LLM, producing $\mathcal{M}^{\text{IT}}$. $\mathcal{M}^{\text{IT}}$ is used during Phase 3 to generate $\set{D}^{\text{IT}}$, that contains both correct and incorrect reasoning trajectories. Phase 4 then performs a weakly supervised LoRA preference optimization with $\set{D}^{\text{IT}}$ to produce the final MATATA model, $\mathcal{M}^{\text{IT+KTO}}$.}
    \label{fig:matata-training}
\end{figure}

\subsubsection{Phase 1: Generation of training dataset.}\label{phase_1} 

During this phase a few-shot prompt-engineered MATATA,
$
\mathcal{M}^{\text{PE}}\left(\cdot \mid \tool{P}^{\text{PE}},\left\{\tool{T}^{\text{PE}}_i\right\}\right)\,
$,
is constructed.
By design, minimal efforts are put into prompt-engineering ($\text{PE}$), as the planner and each tool have to be prompted with in-context-learning examples. 
More details are provided in Section~\ref{sec:experiments} and in Appendix~\ref{appendix-b}.

$\mathcal{M}^{\text{PE}}$ is then applied to the training dataset $\set{D}$ to construct a baseline dataset
\begin{equation}
    \set{D}^\text{PE} = \Bigl\{\left\langle \vec{q}_d,\vec{c}_d,\vec{t}_d, \vec{a}_d, {\trajectory}_{d}\right\rangle \mid \forall d \in \llbracket 1, D \rrbracket\ \text{iff}\ \mathcal{M}^{\text{PE}}\left(\vec{q}_d,\vec{c}_d,\vec{t}_d\right)= \vec{\hat{a}_d}\equiv\vec{a}_d\Bigl\}\,
\end{equation}
that contains all multi-step reasoning trajectories for problems in which $\mathcal{M}^{\text{PE}}$ generates the correct final answers.

\subsubsection{Phase 2: Weakly supervised LoRA instruction tuning.} \label{phase_2}
Relying on the data generated during Phase 1, Instruction Tuning (IT) is applied to train specialized Low-Rank Adaptation (LoRA) \cite{hu2021loralowrankadaptationlarge} adapters for the planner and each of the tools.

For each tool $\tool{T}_i$, a specific IT dataset is extracted from $\set{D}^\text{PE}$ as
\begin{align}
\begin{split}
\set{D}^\text{PE}_i = \Bigl\{ &
    \left\langle
        \tool{input_i}, \tool{output_i}
    \right\rangle_{d}
    \mid
    \forall d \in \set{D}^\text{PE} \text{ s.t. } \tool{T}_i \in {\trajectory}_{d} \\
    &\text{ and } \tool{T}_i(\tool{input_i}) = \tool{output_i} \Bigl\}\,,
\end{split}
\end{align}
where for $\tool{T}_i$ and a given instance $d$, $\tool{input_i} = \left(\vec{q}, \vec{c}_{i-1}, \vec{t}_{i-1}, \vec{a}_{i-1}\right)_d$ is the tool input, and $\tool{output_i} = \left(\vec{q}, \vec{c}_i, \vec{t}_i, \vec{a}_i\right)_d$ is the tool output.


For the planner, its training dataset, $\set{D}^\text{PE}_P$, is expressed as
\begin{equation}
\begin{split}
    \set{D}^\text{PE}_P = & 
        \Bigl\lbrace
            \left\langle 
                (\vec{q}_d,\vec{c}_d,\vec{t}_d), \{\tool{T}_i\}_{\tool{T}_i\in {\trajectory}_{d}}
            \right\rangle
        \mid \forall d \in \llbracket 1, D \rrbracket
        \\
    & \text{iff}\ \mathcal{M}^{\text{PE}}\left(\vec{q}_d,\vec{c}_d,\vec{t}_d\right)\equiv\vec{a}_d
        \Bigl\rbrace\,.
\end{split}
\end{equation}

Training the planner and the tools only relies on the end-to-end weak supervision label, without intermediate supervision or annotation for each tool or the planning. Thus, $ \mathcal{M}^{\text{IT}}\left(\cdot \mid \tool{P}^{\text{IT}},\left\{\tool{T}^{\text{IT}}_i\right\}\right)\,$
is trained through weakly supervised instruction tuning.

\subsubsection{Phase 3: Generation of preference optimization dataset.}

In this phase, the instruction-tuned MATATA, $\mathcal{M}^{\text{IT}}$, obtained in the previous phase, is applied on $\set{D}$ to construct a weakly supervised dataset that will be used in the next phase as part of the Kahneman--Tversky Optimization (KTO) \cite{ethayarajh2024kto} stage.

Consequently, both positive (correct) and negative (incorrect) trajectories are extracted from $\mathcal{M}^{\text{IT}}(\set{D})$ to produce $\set{D}^\text{IT}_{+}$ and $\set{D}^\text{IT}_{-}$,
\begin{align}
    \set{D}^\text{IT}_{+} & = \Bigl\{\left\langle \vec{q}_d,\vec{c}_d,\vec{t}_d, \vec{a}_d, {\trajectory}_{d}\right\rangle \mid \forall d \in \llbracket 1, D \rrbracket\ \text{iff}\ \mathcal{M}^{\text{IT}}\left(\vec{q}_d,\vec{c}_d,\vec{t}_d\right)\equiv\vec{a}_d\Bigl\}\,,\text{ and}\\
    \set{D}^\text{IT}_{-} & = \Bigl\{\left\langle \vec{q}_d,\vec{c}_d,\vec{t}_d, \vec{a}_d, {\trajectory}_{d}\right\rangle \mid \forall d \in \llbracket 1, D \rrbracket\ \text{iff}\ \mathcal{M}^{\text{IT}}\left(\vec{q}_d,\vec{c}_d,\vec{t}_d\right)\neq\vec{a}_d\Bigl\}\,.
\end{align} 

\subsubsection{Phase 4: Weakly supervised LoRA preference optimization.}\label{phase_4} Preference optimization is applied to further fine-tune the specialized LoRA adapter models obtained from Phase 2.
By defining $\set{D}^\text{IT} = \set{D}^\text{IT}_{+}\cup\set{D}^\text{IT}_{-}$, for each tool $\tool{T}_i$ as
\begin{align}
\begin{split}
\set{D}^\text{IT}_i = \Bigl\{ &
    \left\langle
        \tool{input_i}, \tool{output_i}
    \right\rangle_{d}, \tool{label}_d
    \mid
    \forall d \in \set{D}^\text{IT} \text{ s.t. } \tool{T}_i \in {\trajectory}_{d} \\
    &\text{ and } \tool{T}_i(\tool{input_i}) = \tool{output_i} \Bigl\}\,,
\end{split}
\end{align}
where for $\tool{T}_i$ and a given instance $d$, $\tool{input}_i = \left(\vec{q}, \vec{c}_{i-1}, \vec{t}_{i-1}, \vec{a}_{i-1}\right)_d$ is the tool input, $\tool{output}_i = \left(\vec{q}, \vec{c}_i, \vec{t}_i, \vec{a}_i\right)_d$ is the tool output, and the end-to-end weak supervision label $\tool{label}_d$ is expressed as
\begin{equation}
\tool{label}_d =
    \begin{cases} 
        +1, & \text{if } d \in \set{D}^\text{IT}_{+}\,, \\ 
        -1, & \text{if } d \in \set{D}^\text{IT}_{-}\,.
    \end{cases}
\end{equation}

Similarly, for the planner the preference optimization dataset is defined as
\begin{equation}
\begin{split}
    \set{D}^\text{IT}_P & = 
    \Bigl\{\left\langle (\vec{q}_d,\vec{c}_d,\vec{t}_d), \{\tool{T}_i\}_{\tool{T}_i\in {\trajectory}_{d}} \right\rangle, \tool{label}_d \mid\ \forall d\in\set{D}^\text{IT}\Bigl\}\,.
\end{split}
\end{equation}

Then, a MATATA instance,
$\mathcal{M}^{\text{IT+KTO}}$,
is trained by weakly supervised preference optimization, where the weights of $\mathcal{M}^{\text{IT}}$ obtained during Phase 2 are fine-tuned  using $\set{D}^\text{IT}_P$ and $\left\{\set{D}^\text{IT}_i\right\}$.
MATATA is the final result of this two-stage end-to-end weakly supervised training.

Data efficiency is major in this phase, achieved through using all reasoning trajectories, hence learning not only from positive example but also from the negative ones.
In fact, each input only requires either a positive or a negative output, unlike most existing preference optimization approaches which require both \cite{wang2024selftrainingdirectpreferenceoptimization, yang2024weaktostrongreasoning}.
Leveraging all incorrect trajectories could also help tools to learn from areas uncovered during Instruction Tuning as shown in \cite{chen2024advancing}, as an alternative to methods that primarily enhance the framework's existing areas of proficiency.

Each data generation phase also redefines the teacher-student paradigm, as the same model generates the trajectories used for the weak supervision.
Consequently, the model self-improves by learning from its own generated reasoning, which can be more efficient than external reasoning data for training, as shown in \cite{zelikman2022star, huang2022largelanguagemodelsselfimprove}.

\section{Experiments}\label{sec:experiments}

MATATA's modular design enables a unified framework for tabular reasoning, allowing tools to be shared across multiple diverse datasets. Taking that into consideration, experiments are conducted on the FinQA \cite{chen2021finqa}, TAT-QA \cite{zhu2021tatqaquestionansweringbenchmark}, and TabMWP \cite{lu2023dynamicpromptlearningpolicy} datasets.

\subsection{Experimental Protocol}

\newcommand{\best}[1]{\textcolor{docugamigreen}{\textbf{\underline{#1}}}}
\newcommand{\bestsection}[1]{\textcolor{RoyalPurple}{\textbf{{#1}}}}

\begin{table}[h!]  
    \centering
    \caption{Accuracy (Acc.) or Exact-Match (EM) scores across evaluation datasets. Prompt Engineering (PE); Fine-Tuning (FT).
    Best overall scores appear \best{underlined}, best section scores appear in \bestsection{bold}.
    On TAT-QA, scores with an asterisk ($\ast$) are reported on the golden test set (publicly available) and others on the leaderboard test set (private).
    }
    \resizebox{\textwidth}{!}{
    \begin{tabular}{@{}l@{}c@{}c@{\quad}c@{\quad}c@{\quad}c@{}}
        \toprule
         & & & \textbf{FinQA} & \textbf{TAT-QA} & \textbf{TabMWP} \\[-0.56em] 
        \textbf{Framework} & \textbf{Model} & \textbf{Method} & \textbf{Acc.}  & \textbf{EM}    & \textbf{Acc.} \\
        \midrule
        \multicolumn{6}{@{}l}{\textbf{Closed-source models}}\\\midrule
        \multirow{2}{*}{Chameleon \cite{lu2023chameleonplugandplaycompositionalreasoning}} & ChatGPT        & PE & ---     & ---     & 93.28 \\
                                   & GPT-4          & PE & ---     & ---     & \best{98.78} \\
                                   \midrule
        \multirow{2}{*}{EEDP \cite{srivastava2024evaluatingllmsmathematicalreasoning}}      & ChatGPT        & PE & 61.88 & 79.73* & ---  \\
                                   & GPT-4          & PE &          \bestsection{76.05} & \best{88.67*} & ---\\\midrule
        \multirow{2}{*}{PoT \cite{chen2023programthoughtspromptingdisentangling}}                   & SC-Codex   & PE & 68.1  & 70.2*  & 81.8 \\
                                & GPT-4   & PE & 74.0  & ---  & --- \\
        \midrule
        \multicolumn{6}{@{}l}{\textbf{Open-source models}}\\
        \midrule
        FinQANet \cite{chen2021finqa}                  & RoBERTa        & FT& 61.24 & ---     & --- \\\midrule
        TagOP   \cite{zhu2021tatqaquestionansweringbenchmark}                   & RoBERTa        & FT & ---     & 50.1  & --- \\\midrule
        \multirow{2}{*}{TORA \cite{gou2024toratoolintegratedreasoningagent}}      & Code-LLama-34B & FT & ---     & ---     & 70.5 \\
                                   & Llama2-70B     & FT&  ---    & ---     & 74.0 \\\midrule
        \multirow{2}{*}{Husky \cite{kim2024huskyunifiedopensourcelanguage}}     & Llama2-7B      & FT & 20.8  & 42.2*  & 77.6 \\
                                   & Llama3-8B      & FT& 20.9  & 42.3*  & 76.6 \\\midrule
        \multirow{3}{*}{TAT-LLM \cite{zhu2024tatllmspecializedlanguagemodel}}   & Llama2-7B      & FT & 65.13 & 76.4  & --- \\
                                   & Llama2-13B     & FT& 71.93 & 77.5 & --- \\
                                   & Llama2-70B     & FT& 76.81 & \bestsection{81.4} & --- \\\midrule 
        \multirow{2}{*}{\textbf{MATATA (Ours)}} & Phi3-mini 3.8B     & FT& 70.10 & 74.2 & 96.66\\
                                                & Ministral-8B & FT& \best{77.59} & 77.6 & \bestsection{98.13}\\
        \bottomrule
    \end{tabular}}
    \label{table:scores}
\end{table}

As described in Section~\ref{sec:matata-training-phases}, during Phase 1, tools and the planner are few-shot prompted. 
Simple, easily reproducible, broadly applicable prompt choices have been made to prioritize increasing performance through iterative fine-tuning rather than optimizing the prompts.
This approach justifies the use of weak supervision instead of extensive prompt engineering efforts.
In particular, for each in-context-learning prompt, 5 examples from any of the 3 training sets have been drawn. Therefore, the examples and instructions have not been optimized to achieve the best performance for the prompt-engineered MATATA.
The same prompts are also used for all different language models in our experiments. 
Additional details can be found in the appendix.

The training sets are used to generate reasoning trajectories during Phase 1 and Phase 3, to train the planner and the tools.
The best MATATA model is selected with respect to scores on the validation sets.
The test scores are reported in Table~\ref{table:scores} using the relative metric for each dataset, either Accuracy or Exact Match.
All datasets are used following the train/test/validation splits specified in the respective papers.
TAT-QA scores are either reported from the online leaderboard test set (private), or on the golden test set (publicly available) with an asterisk ($\ast$). 
They respectively contain 1,669 and 1,663 questions, with 1,657 in common.
Therefore, scores are first reported from the leaderboard when available, else from their respective papers.
Table~\ref{table:evolution_scores} reports golden test scores over the training stages.

Two light-weight language models are used for prompt-engineering and fine-tuning: microsoft/Phi-3-mini-4k-instruct 3.8B \cite{abdin2024phi3technicalreporthighly} and mistralai/Ministral-8B-Instruct-2410 \cite{ministral2024}.
All trainings were conducted on a single NVIDIA A100 40GB GPU provided Indiana University's Jetstream2 \cite{10.1145/3437359.3465565} under NSF ACCESS \cite{10.1145/3569951.3597559}, and the overall fine-tuning of a MATATA model instance takes on average 16 GPU hours, hence a total of 128 GPU hours for 2 SLMs and 4 instances each (\emph{e.\,g.} Table~\ref{table:evolution_scores}). 
Throughout the training process, one adapter is trained for each tool $\tool{T}_i$ following: $\tool{T}_i^\text{PE} \rightarrow \tool{T}_i^\text{IT} \rightarrow \tool{T}_i^{IT+KTO}$.
During the two-stage training, LoRA 
adapters are continuously fine-tuned from the corresponding baseline language model.
Both Instruction Tuning and KTO preference optimization use the following LoRA hyperparameters: a learning rate of $10^{-5}$, batch size of $32$ over $10$ epochs, with a LoRA $r=64$, $\alpha=32$ and a dropout of $0.05$.
KTO uses the default $\beta$ parameters at $0.1$, and task-specific weights are applied to address class imbalance between valid and invalid trajectories.

\begin{table}[tb]
    \centering
    \caption{Comparison of accuracy or exact match over fine-tuning stages for MATATA-3.8B and MATATA-8B; with one vs. all training set(s). On TAT-QA, scores with an asterisk ($\ast$) are reported on the publicly available golden test set.}
    \begin{tabular}{lc@{\quad}c@{\quad}c@{\quad}c}
        \toprule
        & \textbf{Fine-tuned} & \textbf{FinQA} & \textbf{TAT-QA}           & \textbf{TabMWP}\\[-0.56em] 
        \textbf{Method}  & \textbf{on} & \textbf{Acc.} & \textbf{EM}           & \textbf{Acc.}\\
        \midrule
        Phi3-mini 3.8B PE   &   N/A   & 47.06   & 43.05$^\ast$  & 83.87  \\
        \midrule
        \multirow{2}{*}{\textbf{\text{\ \ \ \ \ }+ weak-sup IT}}    &  One  & 62.00  & 68.19$^\ast$ & 92.89  \\
            &   All  & 67.74  & 71.56$^\ast$  & 96.36  \\
            \midrule
        \multirow{2}{*}{\textbf{\text{\ \ \ \ \ }+ weak-sup KTO}}    &  One  & 69.75& 71.74$^\ast$ & 94.05  \\
             &  All  & \best{70.10} & \best{74.44}$^\ast$ & \best{96.66} \\
        \midrule
        Ministral 8B PE   &   N/A   & 57.11   & 51.29$^\ast$ & 80.89\\\hline
        \multirow{2}{*}{\textbf{\text{\ \ \ \ \ }+ weak-sup IT}}    &  One  & 70.97 & 75.59$^\ast$ & 89.85  \\
            &   All  & 75.68 & 76.19$^\ast$ & 96.64  \\\hline
        \multirow{2}{*}{\textbf{\text{\ \ \ \ \ }+ weak-sup KTO}}    &  One  & 74.63 & 76.36$^\ast$ & 93.38  \\
             &  All  & \best{77.59} & \best{77.81}$^\ast$ & \best{98.13} \\
        \bottomrule
    \end{tabular}
    \label{table:evolution_scores}
\end{table}

\subsection{Experimental Results}

Experimental results (see Table~\ref{table:scores}) show the performance of MATATA-Phi3-mini 3.8B and MATATA-Ministral 8B, respectively referred to as MATATA-3.8B and MATATA-8B. 
Results show that MATATA outperforms Husky, which relies on training data annotated by GPT-4, and PoT-Codex, a GPT-3 variant, on all 3 datasets.
MATATA also surpasses both RoBERTa-based baseline models for FinQA and TAT-QA.

On TAT-QA, MATATA-8B ranges between TAT-LLM-13B and TAT-LLM-70B, while TAT-LLM uses human-expert annotations for training.
Notably, the gap between MATATA-8B and TAT-LLM-70B is only 3.8\%, despite the decrease in model size.
On FinQA, MATATA-8B is above all methods, and MATATA-3.8B ranges between TAT-LLM-7B and 13B, and between ChatGPT and GPT-4 with EEDP-prompting, while being much smaller than those models.
On TabMWP, MATATA-8B exceeds all methods not using GPT-4, ranking 2nd on the TabMWP leaderboard, with only a 0.65\% difference behind Chameleon which relies on the massively larger GPT-4.
For its range of parameters and size, MATATA competes with models fine-tuned on manually annotated reasoning data, or extensively prompted. 
This also confirms the effectiveness of the training method behind MATATA, intuiting that fine-tuning a larger model with weak supervision could further improve those scores.

On TAT-QA with MATATA-8B, the model improvements after weakly supervised instruction tuning are strongest in high-frequency categories (e.g., ’span’ on text and ’arithmetic’ on tables).
However, after preference optimization, underrepresented categories improve more: \textit{count} on tables goes from 64\% to 82\%, and \textit{arithmetic} on text improves from 63\% to 73\%, while well-covered categories show smaller improvements. 
This suggests preference optimization mitigates bias by effectively leveraging both correct and incorrect trajectories from earlier phases.

Two training strategies are explored to investigate knowledge transfer across datasets with shared tools: (i) training the tools on separate datasets, and (ii) training shared tools on the combination of datasets.
The results reported in Table~\ref{table:evolution_scores} show that training tools with weak trajectories from all datasets systematically outperforms single dataset training, in both the IT and KTO stages. 
This improvement can be attributed to better coverage over the trajectories of each tool, with the increase and diversification of training data. 

It also suggests the reusability and scalability of MATATA: the same framework could be enhanced with other datasets and more shared tools, bringing potential improvements to existing results.
Similarly, enhancing MATATA with weakly supervised fine-tuning yields performance improvements ranging from 12.79\% to 31.39\% for the 3.8B model and 17.23\% to 31.39\% for the 8B model.
Although the prompt-engineered model receives minimal initial guidance, applying weak supervision with instruction tuning and preference optimization enables tools to adopt their expected behaviors without intermediary annotations, significantly improving the initial performance.

\section{Conclusion}\label{sec:conclu}

This work introduced MATATA, a novel method for end-to-end weak supervision training for tool-augmented reasoning agents.
MATATA has a focus on tackling tabular data problems, and applies two-stage training to iteratively improve its modules (planner and tool set).
A novel aspect of this approach is that it leverages pretrained open-source SLMs without ever relying on larger or closed-source models or manual annotations.

Thanks to self-improvement \cite{huang2022largelanguagemodelsselfimprove}, MATATA learns from its own generated multi-step reasoning while minimizing both prompt engineering efforts and human supervision.
The experimental results demonstrate that MATATA's reaches state-of-the-art performance in tabular document reasoning, while requiring only a fraction of the resources needed by comparable approaches.

In addition to competitive performance, MATATA also offers:
\begin{itemize}
    \item \textbf{Generalization and adaptability}. One important strength of MATATA is derived from its self-supervised nature. 
    During the iterative training, minimal effort is dedicated to hand-crafting the prompts. 
    This reduces the burden on the user, as the framework adapts to its users, rather than the other way around.
    \item \textbf{Lean, scalable, cost-effective models with reduced computational requirements}.
    The combination of SLMs and an efficient fine-tuning approach reduces the computational footprint of this proposal, both at training and at inference time, offering an alternative perspective on the traditional performance \emph{vs.} parameter laws \cite{DESISLAVOV2023100857}.
    This is particularly relevant for multi-purpose generative architectures such as MATATA, which have been shown to be among the most energy demanding while not necessarily following the \emph{bigger-is-better} assumption \cite{Luccioni_2024}, and with prompt-engineered solutions reported on the higher end of the scale \cite{husom2024pricepromptingprofilingenergy}.
    \item \textbf{Privacy and security when operating on sensitive information}.
    This is essential when handling sensitive business documents like those in FinQA and TAT-QA. 
    Processing those documents with closed-source LLMs hosted off-site by third-party providers can raise security and privacy concerns.
    In contrast, MATATA proposes a novel method to build lightweight and efficient tool-augmented reasoning frameworks, offering the possibility of local deployment for document understanding while ensuring data privacy.

\end{itemize}
This study contributes to the development of fine-tuned multi-step reasoning frameworks for complex tasks, while aiming to minimize overhead and reduce dependencies.

\begin{credits}
\subsubsection{\ackname} 
This work was supported in part by the National Science Foundation SBIR Phase II grant 2233508 ``Authoring Assistance via Contextual Semantic Labeling'' and by the Mathematics of Information Technology and Complex Systems (MITACS) Accelerate grant IT41737. 
This work used Jetstream2 at Indiana University through allocation CIS230177: Assessing Next Generation LLMs for Contextual Semantic Labeling from the Advanced Cyberinfrastructure Coordination Ecosystem: Services \& Support (ACCESS) program, which is supported by U.S. National Science Foundation grants \#2138259, \#2138286, \#2138307, \#2137603, and \#2138296.

\subsubsection{\discintname}
    The authors have no competing interests to declare that are relevant to the content of this article.
\end{credits}

\bibliographystyle{splncs04}
\bibliography{pakdd-2025-matata}

\appendix
\section{Additional Experiment Details}

Additional details of the experiments over training data are summarized in Table \ref{table:training_performances}. 
Correct examples from the prompt-engineered model are used for Instruction Tuning, while correct and incorrect examples from Instruction Tuning are used for KTO preference optimization.

\begin{table}[tb]
    \centering
    \caption{Performance on training data of MATATA-3.8B and MATATA-8B over different configurations of the training sets (in \% points).}
    \begin{tabular}{l@{\quad}lc@{\quad}c@{\quad}ccc}
        \toprule
         &  & \textbf{Training} & \multicolumn{3}{c}{\textbf{Examples} }& \\
        \textbf{Dataset} & \textbf{Model} & \textbf{set} & \textbf{Total} & \textbf{Correct} & \textbf{Incorrect} \\
        \midrule
        \multirow{3}{*}{FinQA} & Phi3-mini 3.8B baseline & N/A & 6251 & 47.87 & 52.13 \\
        & \multirow{2}{*}{+ weak-sup IT} & One & 6251 & 60.09 & 39.91 \\
        & & All & 6251 & \best{66.36} & 33.64 \\\midrule
        \multirow{3}{*}{TabMWP} & Phi3-mini 3.8B baseline & N/A & 23059 & 85.91 & 14.09 \\
        & \multirow{2}{*}{+ weak-sup IT} & One & 23059 & 92.44 & 7.56 \\
        & & All & 23059 & \best{96.62} & 3.38 \\\midrule
        \multirow{3}{*}{TAT-QA} & Phi3-mini 3.8B baseline & N/A & 13205 & 47.58 & 52.42 \\
        & \multirow{2}{*}{+ weak-sup IT} & One & 13205 & 69.08 & 30.92 \\
        & & All & 13205 & \best{75.50} & 24.50 \\\midrule
        \multirow{3}{*}{FinQA} & Ministral 8B baseline & N/A & 6251 & 54.95 & 45.05 \\
        & \multirow{2}{*}{+ weak-sup IT} & One & 6251 & 77.24 & 22.76 \\
        & & All & 6251 & \best{78.40} & 21.60 \\\midrule
        \multirow{3}{*}{TabMWP} & Ministral 8B baseline & N/A & 23059 & 82.49 & 17.51 \\
        & \multirow{2}{*}{+ weak-sup IT} & One & 23059 & 90.60 & 9.40 \\
        & & All & 23059 & \best{97.30} & 2.70 \\\midrule
        \multirow{3}{*}{TAT-QA} & Ministral 8B baseline & N/A & 13205 & 50.78 & 49.22 \\
        & \multirow{2}{*}{+ weak-sup IT} & One & 13205 & 80.60 & 19.40 \\       & & All & 13205 & \best{83.64} & 16.36 \\
        \bottomrule
    \end{tabular}
    \label{table:training_performances}
\end{table}

\section{Prompts Used in Phase 1} \label{appendix-b}

Prompts used in Phase 1 are inspired from Chameleon~\cite{lu2023chameleonplugandplaycompositionalreasoning} and TAT-LLM~\cite{zhu2024tatllmspecializedlanguagemodel}, and adapted for the tasks and tools at hand.
Although, prompt-engineering efforts are minimal: only 5 examples for each tool are given for the in-context learning, and the examples are not modified to optimize the performances of the prompt-engineered model.
Although, prompt-engineering efforts are minimal: only 5 examples for each tool are given for the in-context learning, and the examples are not modified to optimize the performances of the prompt-engineered model.
Moreover, the same prompts are used for both Ministral-8B and Phi-3-3.8B.

\begin{lstlisting}[caption={Prompt for Planner}, label={lst:planner_prompt}]
You need to act as a policy model, that given a question and a modular set, determines the sequence of modules that can be executed sequentially to solve the question.

The modules are defined as follows:

- Row_Lookup: This module returns a simplified table that only remains the rows that are relevant to the question, when the table has more than five rows and can be simplified.

- Column_Lookup: This module returns the simplified table that only remains the columns that are relevant to the question, when the table has more than five columns and can be simplified.

- Context_Extractor: This module extracts relevant information from the context that can be used to answer the question. This is used when the context is too large and can be simplified.

- Span_Extractor: This module extracts the exact span from the context or the table to answer the question. This is used when the question only involves extracting a span of text or numbers.

- Knowledge_Retrieval: This module retrieves domain-specific knowledge for the given question and table. This is used when the question involves domain-specific knowledge, such as function tables, tax forms, etc.

- Program_Generator_And_Verifier: This module generates a python program that can solve the given question. It takes in the question and possible context and produces a program that can be executed by the Program_Executor module. This is used when the question involves computation, such as arithmetic operations over multiple numbers, or logical operations, such as if-else statements.

- Program_Executor: This module executes the generated program from Program_Generator_And_Verifier and produces an output that can be further processed by other modules.

- Solution_Generator: This module generates a detailed solution in natural language to the question based on the information provided. This is used when the question does not require computation.

- Scale_Finder: This module finds the scale associated with the answer when there is one. This is used only if the question asks for the scale of the answer.

- Answer_Generator: This module parses and extracts the final answer from the solution or execution result. It should always be used as the last module in the sequence.

Below are some examples that map problems to modules.
You must only return the sequence of modules separated by spaces.

CONTEXT:
Results of Operations: Years Ended December 31, 2018, versus Year Ended December 31, 2017 (Amounts in thousands, except percentages and per share amounts): Results of Operations: Years Ended December 31, 2018, versus Year Ended December 31, 2017 (Amounts in thousands, except percentages and per share amounts):
Interest expense decreased in the year ended December 31, 2018, versus the same period in 2017 primarily due to lower debt balances, a reduction in interest related to interest rate swaps, and a one-time charge related to a liability that was settled in 2017. 

TABLE:
 | Years Ended December 31, | 
 | 2018 | 2017
Interest expense | $(2,085) | $(3,343)
Interest income | 1,826  | 1,284
Other (expense) income | (2,676) | 3,817
Total other (expense) income, net | $(2,935) | $1,758

QUESTION: What was the percentage change in the net Total other (expense) income between 2017 and 2018?

MODULES: Knowledge_Retrieval Context_Extractor Program_Generator_And_Verifier Program_Executor Answer_Generator
#END

TABLE:
Date | Description | Received | Expenses | Available Funds
 | Balance: end of July | | | $213.05
8/8 | birthday money | $20.00 | | $233.05
8/17 | snow cone | | $3.40 | $229.65
8/21 | backpack | | $27.85 | $201.80

QUESTION: This is Addison complete financial record for August. How much money did Addison spend on her backpack?

MODULES: Column_Lookup Span_Extractor Answer_Generator
END

CONTEXT:
Note 18 Composition of Certain Financial Statement Captions
(1) During the year ended January 3, 2020 and December 28, 2018, the Company recognized $417 million and $146 million, respectively, of amortization related to its transition costs and project assets. (2) Balance represents items that are not individually significant to disclose separately.
(3) Balances are net of $25 million and $29 million of dividends received during fiscal 2019 and fiscal 2018, respectively, that were recorded in cash flows provided by operating activities of continuing operations on the consolidated statements of cash flows.
(4) During the year ended January 3, 2020, the Company combined Dividends payable and Income taxes payable with Accounts payable and accrued liabilities on the consolidated balance sheets. As a result, the prior year activity has been reclassified to conform with the current year presentation.

TABLE:
Balance Sheet | January 3, 2020 | December 28, 2018
 | (in millions) | 
Other current assets:   |  | 
Transition costs and project assets(1) | $98 | $145
Pre-contract costs | 6 | 41
Other(2) | 306 | 357
 | $410 | $543
Other assets: |  | 
Transition costs and project assets(1) | $207 | $22
Equity method investments(3) | 19 | 26
Other(2) | 200 | 134
 | $426 | $182

QUESTION: What was the change in the Equity method investments from 2018 to 2020 ?

MODULES: Row_Lookup Program_Generator_And_Verifier Program_Executor Answer_Generator
#END

CONTEXT:
Company balance sheet
At 31 March 2019
The financial statements were approved by the Board of Directors on 6 June 2019 and authorised for issue.

TABLE:
 |  | 2019 | 2018
 | Note | m | m
Fixed assets |  |  | 
Investments | 3 | 1,216.0 | 1,212.9
 |  | 1,216.0 | 1,212.9
Current assets |  |  | 
Debtors | 4 | 415.9 | 440.7
Cash and cash equivalents | 5 |  | 0.2
 |  | 415.9 | 440.9
Creditors: amounts falling due within one year | 6 | (411.4) | (288.4)
Net current assets |  | 4.5 | 152.5
Net assets |  | 1,220.5 | 1,365.4
Capital and reserves |  |  | 
Called-up share capital | 9 | 9.3 | 9.5
Own shares held | 10 | (16.5) | (16.9)
Capital redemption reserve |  | 0.7 | 0.5
Retained earnings |  | 1,227.0 | 1,372.3
Total equity |  | 1,220.5 | 1,365.4

QUESTION: Who approved the financial statements?

MODULES: Span_Extractor Answer_Generator
#END

CONTEXT:
The Company consolidated net revenues disaggregated by product group are presented in Note 19. The following tables present the Company consolidated net revenues disaggregated by geographical region of shipment and nature.
(1) Net revenues by geographical region of shipment are classified by location of customer invoiced or reclassified by shipment destination in line with customer demand. For example, products ordered by U.S.-based companies to be invoiced to Asia Pacific affiliates are classified as Asia Pacific revenues.
(2) Original Equipment Manufacturers (OEM) are the end-customers to which the Company provides direct marketing application engineering support, while Distribution customers refers to the distributors and representatives that the Company engages to distribute its products around the world.

TABLE:
 |  | Year ended | 
 | December 31, 2019 | December 31, 2018 | December 31, 2017
Net revenues by nature |  |  | 
Revenues from sale of products | 9,381 | 9,461 | 8,175
Revenues from sale of services | 148 | 151 | 133
Other revenues | 27 | 52 | 39
Total revenues | 9,556 | 9,664 | 8,347

QUESTION: What is the average of Other revenues? Provide the scale if there is one.

MODULES: Program_Generator_And_Verifier Program_Executor Scale_Finder Answer_Generator
#END
\end{lstlisting}

\begin{lstlisting}[caption={Prompt for Row Lookup}, label={lst:rl_prompt}]
Read the following question and table. Each row is separated by a newline ('\n') and each column is separated by a vertical bar ('|'). Return the simplified table that only remains the rows that are relevant to the question. If all rows are relevant, or there are less than five rows, return the original table.

QUESTION: Look at the following schedule. Amy is at Oakland. If she wants to arrive at Danville at 3.15 P.M., what time should she get on the train?

TABLE:
Bloomington | 6:30 A.M. | 6:45 A.M. | 8:45 A.M. | 10:30 A.M. | 11:00 A.M.
Oakland | 7:45 A.M. | 8:00 A.M. | 10:00 A.M. | 11:45 A.M. | 12:15 P.M.
Yardley | 9:45 A.M. | 10:00 A.M. | 12:00 P.M. | 1:45 P.M. | 2:15 P.M.
Middletown | 10:15 A.M. | 10:30 A.M. | 12:30 P.M. | 2:15 P.M. | 2:45 P.M.
Danville | 11:15 A.M. | 11:30 A.M. | 1:30 P.M. | 3:15 P.M. | 3:45 P.M.
Castroville | 12:00 P.M. | 12:15 P.M. | 2:15 P.M. | 4:00 P.M. | 4:30 P.M.
Manchester | 1:30 P.M. | 1:45 P.M. | 3:45 P.M. | 5:30 P.M. | 6:00 P.M.

SIMPLIFIED TABLE:
Oakland | 7:45 A.M. | 8:00 A.M. | 10:00 A.M. | 11:45 A.M. | 12:15 P.M.
Danville | 11:15 A.M. | 11:30 A.M. | 1:30 P.M. | 3:15 P.M. | 3:45 P.M.
#END

QUESTION: What is the percentage increase in Retained earnings after adoption of new standard ?

TABLE:
 |  | As of March 29, 2019 | 
(In millions) | As Reported | Balances Without Adoption of New Standard | Effect of Change
Accounts receivable, net | $708 | $657 | $51
Other current assets (1) | $435 | $421 | $14
Other long-term assets (2) | $1,262 | $1,213 | $49
Total assets | $15,938 | $15,824 | $114
Short-term contract liabilities | $2,320 | $2,437 | $(117)
Other current liabilities | $533 | $494 | $39
Long-term contract liabilities | $736 | $837 | $(101)
Deferred income tax liabilities | $577 | $526 | $51
Total liabilities | $10,200 | $10,328 | $(128)
Accumulated other comprehensive loss | $(7) | $(2) | $(5)
Retained earnings | $933 | $686 | $247
Total stockholders' equity | $5,738 | $5,496 | $242

SIMPLIFIED TABLE:
 |  | As of March 29, 2019 | 
(In millions) | As Reported | Balances Without Adoption of New Standard | Effect of Change
Retained earnings | $933 | $686 | $247
#END

TABLE:
balance december 31 2002 | $ 450697000
additions during period 2014 depreciation and amortization expense | 68125000
deductions during period 2014 disposition and retirements of property | -4645000 ( 4645000 )
balance december 31 2003 | 514177000
additions during period 2014 depreciation and amortization expense | 82551000
deductions during period 2014 disposition and retirements of property | -1390000 ( 1390000 )
balance december 31 2004 | 595338000
additions during period 2014 depreciation and amortization expense | 83656000
deductions during period 2014 disposition and retirements of property | -15244000 ( 15244000 )
balance december 31 2005 | $ 663750000

QUESTION: what is the increase observed in the balance at the end of the year during 2005 and 2004?

SIMPLIFIED TABLE:
balance december 31 2002 | $ 450697000
balance december 31 2003 | 514177000
balance december 31 2004 | 595338000
balance december 31 2005 | $ 663750000
#END

QUESTION: In which year was the Lease commitment less than 10,000 thousands?

TABLE:
 | Lease Commitment | Non-Lease Commitment | Total Commitment
 | $ | $ | $
Payments |  |  | 
2020 | 69,617 | 37,089 | 106,706
2021 | 54,195 | 26,948 | 81,143
2022 | 22,978 | 8,189 | 31,167
2023 | 9,227 | - | 9,227
2024 | 5,713 | - | 5,713
Thereafter | - | - | -
Total payments | 161,730 | 72,226 | 233,956
Less: imputed interest | (13,128) |  | 
Carrying value of operating lease liabilities | 148,602 |  | 
Less current portion | (61,431) |  | 
Carrying value of long-term operating lease liabilities | 87,171 |  | 

SIMPLIFIED TABLE:
 | Lease Commitment | Non-Lease Commitment | Total Commitment
 | $ | $ | $
Payments |  |  | 
2020 | 69,617 | 37,089 | 106,706
2021 | 54,195 | 26,948 | 81,143
2022 | 22,978 | 8,189 | 31,167
2023 | 9,227 | - | 9,227
2024 | 5,713 | - | 5,713
#END

QUESTION: In preparation for graduation, some teachers and students volunteered for the various graduation committees. How many people are on the music committee?

TABLE:
Committee | Students | Teachers
Program | 5 | 17
Ticket | 20 | 5
Music | 20 | 15
Schedule | 15 | 20
Food | 18 | 2
Drinks | 10 | 5
Tables | 5 | 10
Chairs | 10 | 15
Decorations | 15 | 20

SIMPLIFIED TABLE:
Committee | Students | Teachers
Music | 20 | 15
#END
\end{lstlisting}

\begin{lstlisting}[caption={Prompt for Column Lookup}, label={lst:cl_prompt}]
Read the following question and table. Each row is separated by a newline ('\n') and each column is separated by a vertical bar ('|'). Return the simplified table that only remains the columns that are relevant to the question. If all columns are relevant, return the original table.

TABLE:
 |  |  |  | As of and for the Year Ended May 31, | 
 (in millions, except per share amounts) | 2019 | 2018 (4) | 2017 (4) | 2016 (4) | 2015 (4)
 | Consolidated Statements of Operations Data: |  |  |  | 
Total revenues | $39,506 | $39,383 | $37,792 | $37,047 | $38,226
Operating income | $13,535 | $13,264 | $12,913 | $12,604 | $13,871
Net income (1) | $11,083 | $3,587 | $9,452 | $8,901 | $9,938
Earnings per share diluted (1) | $2.97 | $0.85 | $2.24 | $2.07 | $2.21
Diluted weighted average common shares outstanding | 3,732 | 4,238 | 4,217 | 4,305 | 4,503
Cash dividends declared per common share | $0.81 | $0.76 | $0.64 | $0.60 | $0.51
 | Consolidated Balance Sheets Data: |  |  |  | 
Working capital (2) | $27,756 | $57,035 | $50,995 | $47,105 | $47,314
Total assets (2) | $108,709 | $137,851 | $136,003 | $112,180 | $110,903
Notes payable and other borrowings (3) | $56,167 | $60,619 | $57,909 | $43,855 | $41,958

QUESTION: Why did the working capital and total assets decrease in fiscal 2019?

SIMPLIFIED TABLE:
As of and for the Year Ended May 31, |  
(in millions, except per share amounts) | 2019 
Consolidated Statements of Operations Data: |  
Total revenues | $39,506
Operating income | $13,535 
Net income (1) | $11,083  
Earnings per share diluted (1) | $2.97 
Diluted weighted average common shares outstanding | 3,732 
Cash dividends declared per common share | $0.81 
Consolidated Balance Sheets Data: |  
Working capital (2) | $27,756
Total assets (2) | $108,709 
Notes payable and other borrowings (3) | $56,167
#END

QUESTION: Amy is at Oakland. If she wants to arrive at Danville at 3.15 P.M., what time should she get on the train?

TABLE:
Oakland | 7:45 A.M. | 8:00 A.M. | 10:00 A.M. | 11:45 A.M. | 12:15 P.M.
Danville | 11:15 A.M. | 11:30 A.M. | 1:30 P.M. | 3:15 P.M. | 3:45 P.M.

SIMPLIFIED TABLE:
Oakland | 11:45 A.M.
Danville | 3:15 P.M.
#END

QUESTION: If current development costs increased in 2008 as much as in 2007, what would the 2008 total be, in millions?

TABLE:
( in millions ) | 2007 | 2006 | 2005
Sales and transfers of oil and gas produced net of production transportation and administrative costs | $ -4887 ( 4887 ) | $ -5312 ( 5312 ) | $ -3754 ( 3754 )
Net changes in prices and production transportation and administrative costs related to future production | 12845 | -1342 ( 1342 ) | 6648
Extensions discoveries and improved recovery less related costs | 1816 | 1290 | 700
Development costs incurred during the period | 1654 | 1251 | 1030
Changes in estimated future development costs | -1727 ( 1727 ) | -527 ( 527 ) | -552 ( 552 )
Revisions of previous quantity estimates | 290 | 1319 | 820 
Net changes in purchases and sales of minerals in place | 23 | 30 | 4557
Accretion of discount | 1726 | 1882 | 1124
Net change in income taxes | -6751 ( 6751 ) | -660 ( 660 ) | -6694 ( 6694 )
Timing and other | -12 ( 12 ) | -14 ( 14 ) | 307
Net change for the year | 4977 | -2083 ( 2083 ) | 4186
Beginning of year | 8518 | 10601 | 6415
End of year | $ 13495 | $ 8518 | $ 1060  
Net change for the year from discontinued operations | $ 2013 | $ -216 ( 216 ) | $ 162 

SIMPLIFIED TABLE:
( in millions ) | 2007  
Sales and transfers of oil and gas produced net of production transportation and administrative costs | $ -4887 ( 4887 )  
Net changes in prices and production transportation and administrative costs related to future production | 12845  
Extensions discoveries and improved recovery less related costs | 1816  
Development costs incurred during the period | 1654  
Changes in estimated future development costs | -1727 ( 1727 )  
Revisions of previous quantity estimates | 290  
Net changes in purchases and sales of minerals in place | 23  
Accretion of discount | 1726  
Net change in income taxes | -6751 ( 6751 )  
Timing and other | -12 ( 12 )  
Net change for the year | 4977  
Beginning of year | 8518  
End of year | $ 13495  
Net change for the year from discontinued operations | $ 2013  
#END

QUESTION: What is the average annual total assets for both Fiscal years?

TABLE:
Year | Non-capital loss carryforwards | Capital loss carryforwards | Undeducted scientific research and development expenses | Depreciation and amortization | Restructuring costs and other reserves | Deferred revenue | Other | Total deferred tax asset | Valuation Allowance | Scientific research and development tax credits | Other (Deferred tax liabilities) | Deferred tax liabilities | Net deferred tax asset | Long-term assets | Long-term liabilities
2019 | $161,119 | 155 | 137,253 | 683,777 | 17,845 | 53,254 | 59,584 | $1,112,987 | $(77,328) | $(14,482) | (72,599) | $(87,081) | $948,578 | 1,004,450 | (55,872)
2018 | $129,436 | 417 | 123,114 | 829,369 | 17,202 | 62,726 | 57,461 | $1,219,725 | $(80,924) | $(13,342) | (82,668) | $(96,010) | $1,042,791 | 1,122,729 | (79,938) 

SIMPLIFIED TABLE:
Year | Net deferred tax asset  
2019 | $948,578
2018 | $1,042,791
#END

QUESTION: Look at the following schedule. When does Recess end?

TABLE:
Subject | Begin | End
Recess | 6:15 A.M. | 7:20 A.M.
Orchestra | 7:30 A.M. | 8:40 A.M.
Art | 8:45 A.M. | 9:35 A.M.
Handwriting | 9:45 A.M. | 10:20 A.M.
Gym | 10:30 A.M. | 11:15 A.M.
Choir | 11:20 A.M. | 12:25 P.M.
Science | 12:35 P.M. | 1:35 P.M.
Reading | 1:40 P.M. | 2:50 P.M.

SIMPLIFIED TABLE:
Subject | End
Recess | 7:20 A.M.
Orchestra | 8:40 A.M.
Art | 9:35 A.M.
Handwriting | 10:20 A.M.
Gym | 11:15 A.M.
Choir | 12:25 P.M.
Science | 1:35 P.M.
Reading | 2:50 P.M.
#END
\end{lstlisting}

\begin{lstlisting}[caption={Prompt for Context Extractor}, label={lst:ce_prompt}]
Read the following table and question, and simplify the context to retrieve what could be helpful for answering the question.

CONTEXT:
Note 21. Equity - dividends
Dividends paid during the financial year were as follows:
The Directors have declared a final dividend of AU 18 cents per share for the year ended 30 June 2019. The dividend will be paid on 25 September 2019 based on a record date of 4 September 2019. This amounts to a total dividend of US$15.9 million based on the number of shares outstanding.
Accounting policy for dividends
Dividends are recognised when declared during the financial year and no longer at the discretion of the company.

QUESTION: What is the price per share for the final dividend for 2019?

SIMPLIFIED CONTEXT:
The Directors have declared a final dividend of AU 18 cents per share for the year ended 30 June 2019.
#END

CONTEXT:
ASSUMPTIONS USED IN STOCK OPTION PRICING MODEL
The fair value of options granted was determined using a variation of a binomial option pricing model that takes into account factors specific to the share incentive plans, such as the vesting period. The following table shows the principal assumptions used in the valuation.

Expected dividend growth is commensurate with BCE's dividend growth strategy. Expected volatility is based on the historical volatility of BCE's share price. The risk-free rate used is equal to the yield available on Government of Canada bonds at the date of grant with a term equal to the expected life of the options

QUESTION: How is the fair value of options granted determined?

SIMPLIFIED CONTEXT:
The fair value of options granted was determined using a variation of a binomial option pricing model that takes into account factors specific to the share incentive plans, such as the vesting period.
#END

CONTEXT:
Results of Operations
Year Ended December 31, 2018 Compared to Year Ended December 31, 2019
During the year ended December 31, 2019, we had an average of 27.2 ships operating in our owned and bareboat fleet (including ships owned by the Partnership), having 9,518 revenue operating days and an average of 27.2 ships operating under our technical management (including 27.0 of our owned and bareboat ships). During the year ended December 31, 2018, we had an average of 26.0 ships operating in our owned and bareboat fleet (including ships owned by the Partnership), having 9,030 revenue operating days and an average of 25.5 ships operating under our technical management (including 25.0 of our owned and bareboat ships).
Revenues: Revenues increased by 8.1%, or $50.3 million, from $618.3 million during the year ended December 31, 2018 to $668.6 million during the year ended December 31, 2019. The increase in revenues is mainly attributable to an increase of $63.4 million deriving from the full operation of the GasLog Houston, the GasLog Hong Kong and the GasLog Gladstone which were delivered on January 8, 2018, March 20, 2018 and March 29, 2018, respectively and the deliveries of the GasLog Gladstone on March 15, 2019 and the GasLog Warsaw on July 31, 2019. These deliveries resulted in an increase in revenue operating days. In addition, there was an increase of $11.0 million from our vessels trading in the spot and short-term market including the impact of the unscheduled dry-dockings of the GasLog Savannah, the GasLog Singapore and the GasLog Chelsea and an increase of $2.7 million from the remaining fleet. The above increases were partially offset by a decrease of $26.1 million from the expiration of the initial time charters of the GasLog Shanghai, the GasLog Santiago, the GasLog Sydney, the GasLog Skagen, the GasLog Saratoga and the Methane Jane Elizabeth and a decrease of $0.7 million due to increased off-hire days from the remaining vessels. The average daily hire rate increased from $68,392 for the year ended December 31, 2018 to $70,167 for the year ended December 31, 2019.

QUESTION: How many ships on average are operating in 2019 and 2018 respectively?

SIMPLIFIED CONTEXT:
During the year ended December 31, 2019, we had an average of 27.2 ships operating in our owned and bareboat fleet (including ships owned by the Partnership), having 9,518 revenue operating days and an average of 27.2 ships operating under our technical management (including 27.0 of our owned and bareboat ships). During the year ended December 31, 2018, we had an average of 26.0 ships operating in our owned and bareboat fleet (including ships owned by the Partnership), having 9,030 revenue operating days and an average of 25.5 ships operating under our technical management (including 25.0 of our owned and bareboat ships).
#END

CONTEXT:
11 Intangible assets (continued)
(a) Intangible assets
RIGHTS AND LICENCES
Certain licences that NEXTDC possesses have an indefinite useful life and are carried at cost less impairment losses and are subject to impairment review at least annually and whenever there is an indication that it may be impaired.
Other licences that NEXTDC acquires are carried at cost less accumulated amortisation and accumulated impairment losses. Amortisation is recognised on a straight-line basis over the estimated useful life. The estimated useful life and amortisation method are reviewed at the end of each annual reporting period.
INTERNALLY GENERATED SOFTWARE
Internally developed software is capitalised at cost less accumulated amortisation. Amortisation is calculated using the straight-line basis over the asset's useful economic life which is generally two to three years. Their useful lives and potential impairment are reviewed at the end of each financial year.
SOFTWARE UNDER DEVELOPMENT
Costs incurred in developing products or systems and costs incurred in acquiring software and licenses that will contribute to future period financial benefits through revenue generation and/or cost reduction are capitalised to software and systems. Costs capitalised include external direct costs of materials and services and employee costs.
Assets in the course of construction include only those costs directly attributable to the development phase and are only recognised following completion of technical feasibility and where the Group has an intention and ability to use the asset.

QUESTION: How was internally developed software capitalised?

SIMPLIFIED CONTEXT:
Internally developed software is capitalised at cost less accumulated amortisation. Amortisation is calculated using the straight-line basis over the asset's useful economic life which is generally two to three years. Their useful lives and potential impairment are reviewed at the end of each financial year.
#END

CONTEXT:
Marketable securities consisted of the following (in thousands):
We classify our marketable securities as available-for-sale. All marketable securities represent the investment of funds available for current operations, notwithstanding their contractual maturities. Such marketable securities are recorded at fair value and unrealized gains and losses are recorded in Accumulated other comprehensive income (loss) until realized.
We typically invest in highly-rated securities with low probabilities of default. Our investment policy requires investments to be rated single A or better, limits the types of acceptable investments, concentration as to security holder and duration of the investment. The gross unrealized gains and losses in fiscal 2019 and 2018 were caused primarily by changes in interest rates.
The longer the duration of marketable securities, the more susceptible they are to changes in market interest rates and bond yields. As yields increase, those securities with a lower yield-at-cost show a mark-to-market unrealized loss. We anticipate recovering the full cost of the securities either as market conditions improve, or as the securities mature. Accordingly, we believe that the unrealized losses are not other-than-temporary.
When evaluating the investments for otherthan- temporary impairment, we review factors such as the length of time and extent to which fair value has been below the amortized cost basis, current market liquidity, interest rate risk, the financial condition of the issuer, and credit rating downgrades. As of December 28, 2019 and December 29, 2018, gross unrealized losses related to our marketable securities portfolio were not material.
QUESTION: Where does the company typically invest?

SIMPLIFIED CONTEXT:
We typically invest in highly-rated securities with low probabilities of default. Our investment policy requires investments to be rated single A or better, limits the types of acceptable investments, concentration as to security holder and duration of the investment. 
#END
\end{lstlisting}

\begin{lstlisting}[caption={Prompt for Span Extractor}, label={lst:se_prompt}]
Read the following tables and questions and then answer extract the span(s) of text(s) that answers the question. Use the context and tables to answer the questions.

CONTEXT:
Company balance sheet
At 31 March 2019
The financial statements were approved by the Board of Directors on 6 June 2019 and authorised for issue.

TABLE:
 |  | 2019 | 2018
 | Note | m | m
Fixed assets |  |  | 
Investments | 3 | 1,216.0 | 1,212.9
 |  | 1,216.0 | 1,212.9
Current assets |  |  | 
Debtors | 4 | 415.9 | 440.7
Cash and cash equivalents | 5 | - | 0.2
 |  | 415.9 | 440.9
Creditors: amounts falling due within one year | 6 | (411.4) | (288.4)
Net current assets |  | 4.5 | 152.5
Net assets |  | 1,220.5 | 1,365.4
Capital and reserves |  |  | 
Called-up share capital | 9 | 9.3 | 9.5
Own shares held | 10 | (16.5) | (16.9)
Capital redemption reserve |  | 0.7 | 0.5
Retained earnings |  | 1,227.0 | 1,372.3
Total equity |  | 1,220.5 | 1,365.4

QUESTION: Who approved the financial statements?

SOLUTION: ['the Board of Directors']
#END

CONTEXT:
There was no material bad debt expense in 2019, 2018 and 2017. In 2019, 2018 and 2017, the Company's largest customer, Apple represented 17.6%, 13.1% and 10.5% of consolidated net revenues, respectively, reported in the ADG, AMS and MDG segments.
In 2019, $75 million of trade accounts receivable were sold without recourse (nil in 2018).

TABLE:
 | December 31, 2019 | December 31, 2018
Trade accounts receivable | 1,396 | 1,292
Allowance for doubtful accounts | (16) | (15)
Total | 1,380 | 1,277

QUESTION: Which is the largest customer of the company?

SOLUTION: ['Apple'].
#END

CONTEXT:
The Company sponsors a defined benefit plan, the Woolworths Group Superannuation Plan (WGSP or the Plan), that provides superannuation benefits for employees upon retirement. The defined benefit plan is closed to new members.

TABLE:
 | 2019 | 2018
 | % | %
Discount rate | 2.9 | 3.8
Expected rate of salary increase | 2.5 | 2.5
Rate of price inflation | 2.0 | 2.0

QUESTION: Is the defined benefit plan open to new members?

SOLUTION: ['The defined benefit plan is closed to new members.']
#END

CONTEXT:
ASSUMPTIONS USED IN STOCK OPTION PRICING MODEL
The fair value of options granted was determined using a variation of a binomial option pricing model that takes into account factors specific to the share incentive plans, such as the vesting period. The following table shows the principal assumptions used in the valuation.
Expected dividend growth is commensurate with BCE's dividend growth strategy. Expected volatility is based on the historical volatility of BCE's share price. The risk-free rate used is equal to the yield available on Government of Canada bonds at the date of grant with a term equal to the expected life of the options

TABLE:
 | 2019 | 2018
Weighted average fair value per option granted | $2.34 | $2.13
Weighted average share price | $58 | $57
Weighted average exercise price | $58 | $56
Expected dividend growth | 5% | 5%
Expected volatility | 14% | 12%
Risk-free interest rate | 2% | 2%
Expected life (years) | 4 | 4

QUESTION: How is the fair value of options granted determined?

SOLUTION: ['using a variation of a binomial option pricing model that takes into account factors specific to the share incentive plans, such as the vesting period']
#END

CONTEXT:
The working capital and total assets decreased in fiscal 2019 primarily due to $36.1 billion of cash used for repurchases of our common stock during fiscal 2019 and also due to dividend payments, partially offset by the favorable impacts to our net current assets resulting from our fiscal 2019 net income.

TABLE:
 |  |  |  | As of and for the Year Ended May 31, | 
 (in millions, except per share amounts) | 2019 | 2018 (4) | 2017 (4) | 2016 (4) | 2015 (4)
 | Consolidated Statements of Operations Data: |  |  |  | 
Total revenues | $39,506 | $39,383 | $37,792 | $37,047 | $38,226
Operating income | $13,535 | $13,264 | $12,913 | $12,604 | $13,871
Net income (1) | $11,083 | $3,587 | $9,452 | $8,901 | $9,938
Earnings per share\u2014diluted (1) | $2.97 | $0.85 | $2.24 | $2.07 | $2.21
Diluted weighted average common shares outstanding | 3,732 | 4,238 | 4,217 | 4,305 | 4,503
Cash dividends declared per common share | $0.81 | $0.76 | $0.64 | $0.60 | $0.51
 | Consolidated Balance Sheets Data: |  |  |  | 
Working capital (2) | $27,756 | $57,035 | $50,995 | $47,105 | $47,314
Total assets (2) | $108,709 | $137,851 | $136,003 | $112,180 | $110,903
Notes payable and other borrowings (3) | $56,167 | $60,619 | $57,909 | $43,855 | $41,958

QUESTION: Why did the working capital and total assets decrease in fiscal 2019?

SOLUTION: ['Working capital and total assets decreased in fiscal 2019 primarily due to $36.1 billion of cash used for repurchases of our common stock during fiscal 2019 and also due to dividend payments, partially offset by the favorable impacts to our net current assets resulting from our fiscal 2019 net income.', 'In addition, our total assets were also affected in all periods presented by the repayments of notes payable and other borrowings']
#END

CONTEXT:
Other income (expense)
Interest income increased $1.2 million primarily as a result of higher weighted-average balances of cash, cash equivalents and investments and higher yields on investments.
Interest expense increased $5.3 million primarily as a result of interest expense of $3.3 million associated with our long-term debt and our financing lease obligation of $2.0 million in connection with the construction of our Lexington, MA \u2013 U.S. headquarters.
Foreign exchange expense and other, net decreased by $3.1 million primarily as a result of a decrease in foreign exchange expense of $1.9 million, sublease income of $0.9 million and a gain on a previously held asset related to the Solebit acquisition of $0.3 million.

TABLE:
Total other income (expense), net | $ (3,781) | $ (2,727) | $ (1,054)

QUESTION: In which year was Total other income (expense), net less than -3,000 thousands?

SOLUTION: ['2018']
#END

CONTEXT:
NOTE 13 - TAXES ON INCOME
B. Deferred income taxes:
Deferred income taxes reflect the net tax effects of temporary differences between the carrying amounts of assets and liabilities for financial reporting purposes and the amounts used for income tax purposes. Significant components of the Company's deferred tax assets are as follows:\nAs of December 31, 2019, the Company has provided a full valuation allowances of $19,911 in respect of deferred tax assets resulting from tax loss carryforward and other temporary differences. Management currently believes that because the Company has a history of losses, it is more likely than not that the deferred tax regarding the loss carryforward and other temporary differences will not be realized in the foreseeable future.

TABLE:
 | December 31 | 
 | 2019 | 2018
 | U.S. $ in thousands | 
Operating loss carryforward | 73,260 | 57,768
Net deferred tax asset before valuation allowance | 19,911 | 15,916
Valuation allowance | (19,911) | (15,916)
Net deferred tax asset | 795 | 772
 
QUESTION: What was the operating loss carryforward amount in 2019 and 2018 respectively?

SOLUTION: ['73,260', '57,768']
#END
\end{lstlisting}

\begin{lstlisting}[caption={Prompt for Solution Generator}, label={lst:sg_prompt}]
Read the following tables and questions and then answer the questions. Use both the context and the table to answer the questions.

CONTEXT:
Stock-based Compensation Expense
The following table sets forth the total stock-based compensation expense resulting from stock options, RSUs, and ESPP included in the Company\u2019s consolidated statements of operations (in thousands):
During the years ended December 31, 2019, 2018, and 2017 the Company capitalized stock-based compensation cost of $0.5 million, $0.1 million, and $0.3 million, respectively, in projects in process as part of property and equipment, net on the accompanying consolidated balance sheets.
As of December 31, 2019, there was $60.3 million unrecognized stock-based compensation expense of which $13.9 million is related to stock options and ESPP and $46.4 million is related to RSUs. The total unrecognized stock-based compensation expense related to stock options and ESPP as of December 31, 2019 will be amortized over a weighted-average period of 2.87 years. The total unrecognized stock-based compensation expense related to RSUs as of December 31, 2019 will be amortized over a weighted-average period of 2.69 years.

TABLE:
 |  | Year Ended December 31, | 
 | 2019 | 2018 | 2017
Cost of revenues | 2,193 | 2,315 | 2,000
Sales and marketing | 6,812 | 6,596 | 6,621
Research and development | 4,804 | 6,137 | 7,949
General and administrative | 18,328 | 16,338 | 15,682
Total stock-based compensation expense | 32,137 | 31,386 | 32,252

QUESTION: How many categories are there under total stock-based compensation expense?

SOLUTION: There are 4 categories under total stock-based compensation expense: stock options, RSUs, ESPP, and total unrecognized stock-based compensation expense.
The answer is 4.
#END

CONTEXT:
Stock-based Compensation Expense
The following table sets forth the total stock-based compensation expense resulting from stock options, RSUs, and ESPP included in the Company\u2019s consolidated statements of operations (in thousands):\nDuring the years ended December 31, 2019, 2018, and 2017 the Company capitalized stock-based compensation cost of $0.5 million, $0.1 million, and $0.3 million, respectively, in projects in process as part of property and equipment, net on the accompanying consolidated balance sheets.
As of December 31, 2019, there was $60.3 million unrecognized stock-based compensation expense of which $13.9 million is related to stock options and ESPP and $46.4 million is related to RSUs. The total unrecognized stock-based compensation expense related to stock options and ESPP as of December 31, 2019 will be amortized over a weighted-average period of 2.87 years. The total unrecognized stock-based compensation expense related to RSUs as of December 31, 2019 will be amortized over a weighted-average period of 2.69 years.

TABLE:
 |  | Year Ended December 31, | 
 | 2019 | 2018 | 2017
Cost of revenues | 2,193 | 2,315 | 2,000
Sales and marketing | 6,812 | 6,596 | 6,621
Research and development | 4,804 | 6,137 | 7,949
General and administrative | 18,328 | 16,338 | 15,682
Total stock-based compensation expense | 32,137 | 31,386 | 32,252

QUESTION: From 2017 to 2019, how many of the years was the research and development more than 5 million?

SOLUTION: In 2017 and 2018, the research and development was more than 5 million. The answer is 2.
#END

TABLE:
Price | Quantity demanded | Quantity supplied
$895 | 21,000 | 3,400
$945 | 17,200 | 7,400
$995 | 13,400 | 11,400
$1,045 | 9,600 | 15,400
$1,095 | 5,800 | 19,400

QUESTION: At a price of $995, is there a shortage or a surplus? 

SOLUTION: At the price of $995, the quantity demanded is greater than the quantity supplied. There is not enough of the good or service for sale at that price. So, there is a shortage.
The answer is shortage.
#END

Read the following table regarding Ferry fares and then answer a question:

TABLE:
Ferry | Bicycle | Car
Mukilteu-Clinton | $5 | $7
Seattle-Bremerton | $8 | $12
Ocracoke | $3 | $15
Southport-Fort Fisher | $2 | $5
Fauntleroy-Vashon | $5 | $15

QUESTION: For an economics project, Santiago determined the cost of ferry rides for bicycles and cars. Of the ferries shown, which charges the least for a bicycle?

SOLUTION: Look at the numbers in the Bicycle column. Find the least number in this column. 
The least number is $2.00, which is in the Southport-Fort Fisher row. The Southport-Fort Fisher ferry charges the least for a bicycle.
The answer is Southport-Fort Fisher.
#END

CONTEXT:
Stock-based Compensation Expense
The following table sets forth the total stock-based compensation expense resulting from stock options, RSUs, and ESPP included in the Company\u2019s consolidated statements of operations (in thousands):\nDuring the years ended December 31, 2019, 2018, and 2017 the Company capitalized stock-based compensation cost of $0.5 million, $0.1 million, and $0.3 million, respectively, in projects in process as part of property and equipment, net on the accompanying consolidated balance sheets.
As of December 31, 2019, there was $60.3 million unrecognized stock-based compensation expense of which $13.9 million is related to stock options and ESPP and $46.4 million is related to RSUs. The total unrecognized stock-based compensation expense related to stock options and ESPP as of December 31, 2019 will be amortized over a weighted-average period of 2.87 years. The total unrecognized stock-based compensation expense related to RSUs as of December 31, 2019 will be amortized over a weighted-average period of 2.69 years.

TABLE:
 |  | Year Ended December 31, | 
 | 2019 | 2018 | 2017
Cost of revenues | 2,193 | 2,315 | 2,000
Sales and marketing | 6,812 | 6,596 | 6,621
Research and development | 4,804 | 6,137 | 7,949
General and administrative | 18,328 | 16,338 | 15,682
Total stock-based compensation expense | 32,137 | 31,386 | 32,252

QUESTION: From 2017 to 2019, how many of the years was the research and development more than 5 million?

SOLUTION: In 2017 and 2018, the research and development was more than 5 million. The answer is 2.
#END

CONTEXT:
The following table shows the fair value of the DB pension plan assets for each category.
Equity securities included approximately $15 million of BCE common shares, or 0.06% of total plan assets, at December 31, 2019 and approximately $8 million of BCE common shares, or 0.03% of total plan assets, at December 31, 2018.
Debt securities included approximately $53 million of Bell Canada debentures, or 0.21% of total plan assets, at December 31, 2019 and approximately $68 million of Bell Canada debentures, or 0.30% of total plan assets, at December 31, 2018.
Alternative investments included an investment in MLSE of $135 million, or 0.53% of total plan assets, at December 31, 2019 and $135 million, or 0.59% of total plan assets, at December 31, 2018.
The Bell Canada pension plan has an investment arrangement which hedges part of its exposure to potential increases in longevity, which covers approximately $4 billion of post-employment benefit obligations.
The fair value of the arrangement is included within other alternative investments. As a hedging arrangement of the pension plan, the transaction requires no cash contributions from BCE.

TABLE:
FOR THE YEAR ENDED DECEMBER 31 | 2019 | 2018
Observable markets data |  | 
Equity securities |  | 
Canadian | 1,017 | 844
Foreign | 4,534 | 3,770
Debt securities |  | 
Canadian | 13,216 | 12,457
Foreign | 2,385 | 2,004
Money market | 219 | 327
Non-observable markets inputs |  | 
Alternative investments |  | 
Private equities | 2,119 | 1,804
Hedge funds | 1,001 | 1,014
Real estate | 948 | 758
Other | 91 | 93
Total | 25,530 | 23,071

QUESTION: How many components are there under alternative investments?

SOLUTION: There are 4 components under alternative investments: private equities, hedge funds, real estate, and other. The answer is 4.
#END
\end{lstlisting}

\begin{lstlisting}[caption={Prompt for Knowledge Retrieval}, label={lst:kr_prompt}]
Read the following table and question, and generate the domain-specific knowledge as the context information that could be helpful for answering the question.

TABLE:
x | y
10 | 15
11 | 9
12 | 2

QUESTION: The table shows a function. Is the function linear or nonlinear?

KNOWLEDGE:
- A linear function is a function whose graph is a straight line.
- A nonlinear function is a function whose graph is not a straight line.
- The equation of a linear function is y = mx + b, where m is the slope and b is the y-intercept.
- The equation of a nonlinear function is not y = mx + b.
#END

TABLE: Dog | Weight change (oz.)
Sprinkles | 5
Champ | -6

QUESTION: Austen has two dogs, Sprinkles and Champ. He is concerned because Sprinkles keeps eating Champ's food. Austen asks their vet how much each dog's weight has changed since their last visit. Which dog's weight has changed the most?

KNOWLEDGE:
- This table shows the weight change (in ounces) for two dogs, Sprinkles and Champ.
- The dog whose weight has changed the most is the one with the highest (positive or negative) weight change.
#END

TABLE:
Stem | Leaf 
1 | 6, 7, 8
2 | 1, 4, 6, 7
3 | 
4 | 1, 2, 5, 7, 7, 9
5 | 1, 8, 9

QUESTION: Ms. Bradford reported her students' scores on the most recent quiz. How many students scored exactly 17 points?

KNOWLEDGE:
- This is a stem-leaf plot, where each data value is split into a "stem" (the first digit or digits) and a "leaf" (usually the last digit).
- The stems represent the tens digit of the data values, while the leaves represent the ones digit.
- The data value 17 would be represented as stem 1 and leaf 7.
#END

TABLE: 
 | Fly | Read minds
Forgetful | 2 | 4
Lazy | 2 | 2

QUESTION: A creative writing class compiled a list of their favorite superheroes. They listed each superhero's superpower and personality flaw. What is the probability that a randomly selected superhero is lazy and can read minds? Simplify any fractions.

KNOWLEDGE:
- This table shows a two-way frequency table, which is used to show the relationship between two variables.
- The probability of an event is calculated by dividing the number of favorable outcomes by the total number of outcomes.
#END

TABLE: 
Stem | Leaf 
3 | 3
4 | 1, 4
5 | 6, 7
6 | 6
7 | 2

QUESTION: The receptionist at a doctor's office kept track of each patient's wait time. What is the shortest wait time?

KNOWLEDGE:
- This is a stem-leaf plot, which is used to organize numerical data. 
- The stems represent the tens digit of the data values, while the leaves represent the ones digit.
- The shortest wait time is represented by the stem with the lowest value and the leaf with the lowest value within that stem.
#END
\end{lstlisting}

\begin{lstlisting}[caption={Prompt for Program Generator}, label={lst:pg_prompt}]
Read the following tables and questions and then write Python code to answer the questions. Sometimes, the context is not useful to solve the question, and sometimes the table is not useful to solve the question. In such cases, you can ignore the context or table.

CONTEXT:
federal realty investment trust schedule iii summary of real estate and accumulated depreciation 2014 continued three years ended december 31 , 2005 reconciliation of accumulated depreciation and amortization .

TABLE:
balance december 31 2002 | $ 450697000
additions during period 2014 depreciation and amortization expense | 68125000
deductions during period 2014 disposition and retirements of property | -4645000 ( 4645000 )
balance december 31 2003 | 514177000
additions during period 2014 depreciation and amortization expense | 82551000
deductions during period 2014 disposition and retirements of property | -1390000 ( 1390000 )
balance december 31 2004 | 595338000
additions during period 2014 depreciation and amortization expense | 83656000
deductions during period 2014 disposition and retirements of property | -15244000 ( 15244000 )
balance december 31 2005 | $ 663750000

QUESTION: what is the increase observed in the balance at the end of the year during 2005 and 2004?

ANSWER:
#Python Code, return 'ans'.
balance_2005 = 663750000
balance_2004 = 595338000
ans = (balance_2005 - balance_2004) / balance_2004 * 100
#END

TABLE:
Name | Number of coins
Shelby | 81
Oliver | 84
Jamal | 78
Vince | 81
Abby | 79
Farid | 77
Tara | 85
Krysta | 83

QUESTION: Some friends discussed the sizes of their coin collections. What is the mean of the numbers? Please reason step by step.

ANSWER:
#Python Code, return 'ans'.
number_of_coins_for_different_person = [81, 84, 78, 81, 79, 77, 85, 83]
ans = sum(number_of_coins_for_different_person) / len(number_of_coins_for_different_person)
#END

CONTEXT:
Advertising Costs: Advertising costs amounted to $278,057, $365,859, and $378,217, for the years ended September 30, 2019, 2018, and 2017, respectively, and are charged to expense when incurred.
Net Income Per Share: Basic and diluted net income per share is computed by dividing net income by the weighted average number of common shares outstanding and the weighted average number of dilutive shares outstanding, respectively.
There were 268,000 and 108,000 shares for the years ended September 30, 2019 and 2018, respectively, that were excluded from the above calculation as they were considered antidilutive in nature. No shares were considered antidilutive for the year ended September 30, 2017.

TABLE:
 | Year ended September 30, |  | 
 | 2019 | 2018 | 2017
Net income | $4,566,156 | $4,274,547 | $3,847,839
Weighted average common shares | 13,442,871 | 13,429,232 | 13,532,375
Dilutive potential common shares | 8,343 | 23,628 | 128,431
Weighted average dilutive common shares outstanding | 13,451,214 | 13,452,860 | 13,660,806
Earnings per share: |  |  | 
Basic | $0.34 | $0.32 | $0.28
Diluted | $0.34 | $0.32 | $0.28

QUESTION: What is the total earnings in 2019 ?

ANSWER:
#Python Code, return 'ans'.
weighted_avg_common_shares = 13442871
basic_earning_per_share = 0.34
ans = basic_earning_per_share*weighted_avg_common_shares
#END

QUESTION: Martin class recorded how many cans of food each student collected for their canned food drive. What is the median of the numbers? Please reason step by step.

ANSWER:
#Python Code, return 'ans'.
cans = [18, 11, 24, 4, 22, 18, 6]
cans = sorted(cans)
middle1 = (len(cans) - 1) // 2
middle2 = len(cans) // 2
ans = (cans[middle1] + cans[middle2]) / 2
#END

CONTEXT:
item 6 . selected financial data the following table sets forth our selected financial data . the table should be read in conjunction with item 7 and item 8 of this annual report on form 10-k. .( 1 ) long-term debt does not include the current portion of long-term debt , which is included in current liabilities . ( 2 ) free cash flow is a non-gaap financial measure and represents cash from operating activities less capital expenditures net of related grant proceeds . see liquidity and capital resources in item 7 for more information on this measure. .

TABLE:
( $ in millions except per share amounts ) | year ended december 31 2017 | year ended december 31 2016 | year ended december 31 2015 | year ended december 31 2014 | year ended december 31 2013
sales and service revenues | $ 7441 | $ 7068 | $ 7020 | $ 6957 | $ 6820
goodwill impairment | 2014 | 2014 | 75 | 47 | 2014
operating income ( loss ) | 865 | 858 | 769 | 655 | 512
net earnings ( loss ) | 479 | 573 | 404 | 338 | 261
total assets | 6374 | 6352 | 6024 | 6239 | 6190
long-term debt ( 1 ) | 1279 | 1278 | 1273 | 1562 | 1665
total long-term obligations | 3225 | 3356 | 3260 | 3562 | 3277
net cash provided by ( used in ) operating activities | 814 | 822 | 861 | 755 | 260
free cash flow ( 2 ) | 453 | 537 | 673 | 590 | 121
dividends declared per share | $ 2.52 | $ 2.10 | $ 1.70 | $ 1.00 | $ 0.50
basic earnings ( loss ) per share | $ 10.48 | $ 12.24 | $ 8.43 | $ 6.93 | $ 5.25
diluted earnings ( loss ) per share | $ 10.46 | $ 12.14 | $ 8.36 | $ 6.86 | $ 5.18

QUESTION: what was the return on total assets during 2013?

ANSWER:
#Python Code, return 'ans'.
net_income = 261
total_assets = 6190 
ans = net_income / total_assets * 100
\end{lstlisting}

\begin{lstlisting}[caption={Prompt for Scale Finder}, label={lst:sf_prompt}]
Read the following tables, questions and solutions and then find the scale associated to the answer when there is one. The scale can only be 'thousand', 'million', 'billion', 'percent': if not, use an empty string. If there is no scale, use an empty string.

CONTEXT:
The Company's consolidated net revenues disaggregated by product group are presented in Note 19. The following tables present the Company's consolidated net revenues disaggregated by geographical region of shipment and nature.
(1) Net revenues by geographical region of shipment are classified by location of customer invoiced or reclassified by shipment destination in line with customer demand. For example, products ordered by U.S.-based companies to be invoiced to Asia Pacific affiliates are classified as Asia Pacific revenues.
(2) Original Equipment Manufacturers ("OEM") are the end-customers to which the Company provides direct marketing application engineering support, while Distribution customers refers to the distributors and representatives that the Company engages to distribute its products around the world.

TABLE:
 |  | Year ended | 
 | December 31, 2019 | December 31, 2018 | December 31, 2017
Net revenues by nature |  |  | 
Revenues from sale of products | 9,381 | 9,461 | 8,175
Revenues from sale of services | 148 | 151 | 133
Other revenues | 27 | 52 | 39
Total revenues | 9,556 | 9,664 | 8,347

QUESTION: What is the average of Other revenues?

SOLUTION: 39.333333333333336

SCALE: ''
#END    

CONTEXT:
Note 18-Composition of Certain Financial Statement Captions
(1) During the year ended January 3, 2020 and December 28, 2018, the Company recognized $417 million and $146 million, respectively, of amortization related to its transition costs and project assets. (2) Balance represents items that are not individually significant to disclose separately.
(3) Balances are net of $25 million and $29 million of dividends received during fiscal 2019 and fiscal 2018, respectively, that were recorded in cash flows provided by operating activities of continuing operations on the consolidated statements of cash flows.
(4) During the year ended January 3, 2020, the Company combined "Dividends payable and "Income taxes payable" with "Accounts payable and accrued liabilities" on the consolidated balance sheets. As a result, the prior year activity has been reclassified to conform with the current year presentation.

TABLE:
Balance Sheet | January 3, 2020 | December 28, 2018
 | (in millions) | 
Other current assets:   |  | 
Transition costs and project assets(1) | $98 | $145
Pre-contract costs | 6 | 41
Other(2) | 306 | 357
 | $410 | $543
Other assets: |  | 
Transition costs and project assets(1) | $207 | $22
Equity method investments(3) | 19 | 26
Other(2) | 200 | 134
 | $426 | $182

QUESTION: What was the change in the Equity method investments from 2018 to 2020 ?

SOLUTION: -7

SCALE: 'million'
#END

CONTEXT:
The following table presents unaudited supplemental pro forma results for fiscal 2019 and 2018 as if both the Grakon acquisition had occurred as of the beginning of fiscal 2018 and the Pacific Insight acquisition had occurred as of the beginning of fiscal 2017. The unaudited pro forma information is presented for information purposes only and is not indicative of the results of operations that would have been achieved if the acquisitions had taken place at such times.
The unaudited pro forma results presented below primarily include amortization charges for acquired intangible assets, depreciation adjustments for property, plant and equipment that has been revalued, interest expense adjustments due to an increased debt level, adjustments for certain acquisition-related charges and related tax effects.

TABLE:
 | Fiscal Year Ended | 
(Dollars in Millions) | April 27, 2019 | April 28, 2018
Revenues | $1,073.3 | $1,095.0
Net Income | $106.4 | $70.5

QUESTION: In which year was net income less than 100.0 million?

SOLUTION: 2018

SCALE: ''
#END

CONTEXT:
The Company recognizes these compensation costs on a straight-line basis over the requisite service period of the award, which is generally the award vesting term of four years. Forfeitures are accounted for as they occur.
Total stock-based compensation cost capitalized in inventory was less than $0.8 million in the years ended December 31, 2019, 2018 and 2017.

TABLE:
 |  | Year Ended December 31, | 
 | 2019 | 2018 | 2017
 |  | (In thousands) | 
Cost of revenue | $2,843 | $2,435 | $1,406
Research and development | 6,532 | 4,283 | 2,968
Sales and marketing | 9,069 | 8,267 | 5,481
General and administrative | 10,693 | 11,476 | 9,114
Total | $29,137 | $26,461 | $18,969

QUESTION: Which years was the total stock-based compensation cost capitalized in inventory less than $0.8 million?

SOLUTION: ['2019', '2018', '2017']

SCALE: ''
#END

CONTEXT:
Results of Operations: Years Ended December 31, 2018, versus Year Ended December 31, 2017 (Amounts in thousands, except percentages and per share amounts): Results of Operations: Years Ended December 31, 2018, versus Year Ended December 31, 2017 (Amounts in thousands, except percentages and per share amounts):
Interest expense decreased in the year ended December 31, 2018, versus the same period in 2017 primarily due to lower debt balances, a reduction in interest related to interest rate swaps, and a one-time charge related to a liability that was settled in 2017. 

TABLE:
 | Years Ended December 31, | 
 | 2018 | 2017
Interest expense | $(2,085) | $(3,343)
Interest income | 1,826 | 1,284
Other (expense) income | (2,676) | 3,817
Total other (expense) income, net | $(2,935) | $1,758

QUESTION: What was the percentage change in the net Total other (expense) income between 2017 and 2018?

SOLUTION: -266.95108077360635

SCALE: percent
#END
\end{lstlisting}

\end{document}